%% file: main.tex
\documentclass{article}

\usepackage{iclr2027_conference,times}
\input{math_commands.tex}

\usepackage{amsmath}
\usepackage{amssymb}
\usepackage{booktabs}
\usepackage{array}
\usepackage{float}
\usepackage{multirow}
\usepackage{tabularx}
\usepackage{placeins}
\usepackage{xcolor}
\usepackage{graphicx}
\usepackage{tikz}
\usetikzlibrary{arrows.meta}
\usepackage{hyperref}
\usepackage{url}
\newcommand{\ghlogo}{\raisebox{-1.5pt}{\includegraphics[height=1.05em]{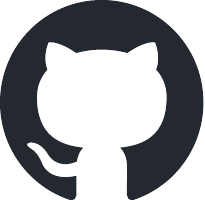}}}
\newcommand{\weblogo}{\raisebox{-1.5pt}{\tikz[x=1.05em,y=1.05em]{%
  \fill (0.5,0.5) circle (0.5);
  \begin{scope}[white,line width=0.45pt,shift={(0.5,0.5)}]
    \draw (0,0) circle (0.33); \draw (0,0) ellipse (0.13 and 0.33);
    \draw (-0.33,0)--(0.33,0); \draw (-0.29,0.16)--(0.29,0.16); \draw (-0.29,-0.16)--(0.29,-0.16);
  \end{scope}}}}

\hypersetup{hidelinks}

\newcommand{\gsw}{\textsc{GSW}}
\newcommand{\paninidr}{\textsc{PANINI-DR}}

\title{\paninidr: Controlled Deep Research\\
through Generative Semantic Workspaces}
\author{M.~Yigit Turali$^{*}$\hspace{2.2em}Shreyas Rajesh$^{*}$\hspace{2.2em}M.~Efe Lorasdagi\hspace{2.2em}Chenda Duan \\[6pt]
\textbf{Kartik Sharma\hspace{2.2em}Vwani Roychowdhury} \\[6pt]
University of California, Los Angeles \\
$^{*}$Equal contribution \\
\texttt{myigitturali@ucla.edu}}
\iclrfinalcopy  %

\begin{document}
\maketitle
\lhead{Under review as a conference paper at ICLR 2027}  %

\begin{abstract}
Deep research requires connecting multiple steps of search, reasoning,
and computation, with intermediate findings guiding what to investigate
next. As investigations grow, agents must maintain research state across
model calls and check findings before errors propagate into subsequent
work. We introduce \paninidr{}, a multi-agent harness that separates
coordination of the overall investigation from the research and
verification of individual subquestions. A shared workspace, built on the
Generative Semantic Workspace representation, records subquestions,
their dependencies, findings, and supporting evidence. Researchers
investigate individual subquestions, while a separate verification agent
checks proposed answers against those subquestions and their evidence.
Verification feedback guides further investigation, and recorded findings
inform subsequent assignments. We evaluate \paninidr{} on BrowseComp-Plus,
FRAMES, and MoNaCo against seven state-of-the-art deep-research harnesses, all using a shared
language-model backbone and frozen corpora. \paninidr{} achieves the
strongest average performance across the three benchmarks. On BrowseComp-Plus,
\paninidr{} achieves \textbf{73.3\%} accuracy,
outperforming the next-best harness by \textbf{16.0} percentage points. Its advantage is strongest on
hard questions, where it answers nearly twice as many as the next-best harness.
Our ablation studies further show that
structured planning and persistent state improve performance, as does
checking each finding against its cited evidence before it is reused.
Because a model can answer correctly from memory alone, we also analyze the
evidence retained by each harness. \paninidr{} achieves the highest answer
recoverability from saved evidence on all three benchmarks, while
complementary claim-support audits show stronger evidence-backed answers.
Accounting for this evidence changes how harnesses compare and further
strengthens \paninidr{}'s advantage.
\end{abstract}

\begin{center}
\href{https://github.com/roychowdhuryresearch/panini-dr}{\ghlogo\,\,\textbf{Code}}\hspace{3em}%
\href{https://roychowdhuryresearch.github.io/PANINI-DR_Project_Page/}{\weblogo\,\,\textbf{Project page}}
\end{center}

\section{Introduction}

Answering questions over collections of documents is one of the most useful
applications of language models. A common early approach was a simple
retrieve-then-answer pipeline: a retriever selected passages relevant to a
question, and a language model used those passages to generate an answer
\citep{lewis2020rag,izacard2021leveraging}. As language models improved at
reasoning and using tools, they became capable of answering increasingly
complex questions \citep{yao2023react,trivedi2023ircot}. One such class is deep
research questions, which require gathering and connecting evidence across
multiple sources \citep{zheng2025deepresearcher,li2025webthinker}. For these
questions, the relevant documents may not be apparent at the outset, making
them difficult to resolve with a single retrieve-then-answer step.
One must first break down the problem into appropriate subquestions, search
for evidence to answer them, and combine the findings into a final answer.
This process can be iterative: answering one subquestion may reveal what
to investigate next, requiring successive rounds of search and reasoning
\citep{press2023selfask,trivedi2023ircot}.

Deep research agents carry out this process through repeated search,
reading, and reasoning over source material
\citep{zheng2025deepresearcher,li2025webthinker}. Consider the following
question, taken from the FRAMES dataset \citep{krishna2024frames}:
``How old was the U.S. President's wife
when the Camp David Accords were signed?'' An agent can begin by establishing
when the accords were signed and who was president at the time. Identifying
\emph{Jimmy Carter} then provides a starting point for finding his wife,
\emph{Rosalynn Carter}, and her birth date. Answering the question then
requires reasoning over the retrieved facts: the agent must combine her
birth date with the signing date to compute her age, accounting for whether
her birthday had occurred that year.

Answering these questions thus requires connecting multiple steps of search,
reasoning, and computation, with intermediate findings guiding what to
investigate or calculate next. Over a longer investigation, these findings
accumulate across multiple searches, model calls, and reasoning steps.
Early approaches to agentic question answering, such as ReAct, used a
single agent to decide what to investigate, invoke tools, and reason over
the results, with progress retained in a growing interaction history
\citep{yao2023react}. When the state of the investigation is left implicit
in this history, the model must recover earlier findings while carrying out further search,
reasoning, and computation. As that history grows, relevant findings can
become harder to use: long-context studies report difficulties using
information in the middle of an input \citep{liu2024lostmiddle} and declining
performance as input length increases \citep{hong2025contextrot}.

These difficulties motivate a separation of concerns between
\textbf{maintaining research state} and performing the search, reasoning,
and computation needed to answer the question. The research state records the subquestions needed to
answer the original query, their dependencies, the findings and supporting
evidence obtained so far, and what remains unresolved. For example,
identifying \emph{Jimmy Carter} as the president resolves one subquestion
and provides the context for the next (identifying his wife at the time).
The need to track progress and dependencies also arises as LLM agents take
on increasingly ambitious tasks in other domains, such as mathematical
formalization. For example, Anthropic reports that early attempts to
formalize Fermat's Last Theorem lost track of project state and struggled
to coordinate. The successful effort followed a switch to Prove2Me, whose
shared graph of theorem statements and
dependencies helped agents maintain progress and coordinate parallel work
\citep{anthropic2026fermat,chen2026prove2me}.
Within deep research, several methods similarly maintain explicit records
of research progress. IterResearch reconstructs the
context at each step using an evolving research report
\citep{chen2025iterresearch}, while SLEUTH organizes research progress into
facts, hypotheses, and open questions \citep{liu2026sleuth}. Dossier
maintains a persistent research ledger of claims, contradictions, and
information gaps, using it to guide parallel search branches and reconcile
their findings \citep{chabra2026dossier}.

The dependencies between subquestions also allow errors to propagate.
In the Camp David example, suppose the agent incorrectly identifies
\emph{Ronald Reagan} rather than \emph{Jimmy Carter} as the president when
the accords were signed. It could then retrieve accurate information about
the wrong person's wife and correctly calculate her age, yet still produce
an incorrect answer.
\textbf{Verifying intermediate findings} against their supporting evidence
and the questions they are intended to resolve provides an opportunity to
identify and correct such errors before they guide subsequent work.
Recent work reports substantial improvements in research-agent performance
from verification feedback. DeepVerifier uses rubric-guided verification
to provide feedback for further attempts \citep{wan2026deepverifier},
while Marco DeepResearch verifies candidate answers to guide continued
research \citep{zhu2026marco}. Other methods integrate verification
directly into the research state: Omni-Decision records evidence needs,
dependencies, and critic judgments, and uses them to guide subsequent
actions \citep{ma2026omnidecision}.

Building on these two ideas (maintaining research state and verifying
intermediate findings) and recent work, we introduce
\textbf{\paninidr{}}, a multi-agent harness for deep research. \paninidr{} maintains a shared
research workspace and coordinates separate agents to investigate
subquestions and verify their proposed answers. The workspace uses the
Generative Semantic Workspace (\gsw{}) representation to express subquestions
and their relationships \citep{rajesh2026gsw,rajesh2026panini}. A central agent uses the
workspace to track overall progress and launch research assignments.
Within each assignment, verification feedback guides further investigation.
The workspace records the resulting findings, supporting evidence, and
review outcomes, making them available to subsequent assignments. This enables us to
allocate research and verification effort to individual subquestions while
preserving their dependencies within the overall investigation.

We evaluate \paninidr{} on BrowseComp-Plus, FRAMES, and MoNaCo, comparing
it with existing state-of-the-art deep research approaches. Among the
evaluated systems, it achieves the strongest average performance across
benchmarks, with strong gains on the hardest questions.
This pattern is consistent with our motivation for maintaining research
state across extended investigations.

To further assess whether correct answers are supported by the research
performed, we analyze the evidence retained by each harness. A model can
answer correctly from its parametric knowledge even when its retrieved
evidence is insufficient.

We summarize our contributions as follows:
(1) \textbf{A shared workspace for research and verification.}
We introduce \paninidr{}, a multi-agent harness that organizes subquestions,
their dependencies, and supporting evidence in a shared workspace. This
separates the coordination of the overall investigation from the research
and verification of individual subquestions.
(2) \textbf{Strong performance supported by ablation studies.}
We show that \paninidr{} achieves the strongest average performance across
the evaluated benchmarks, with particularly strong gains on harder
questions. Through ablation studies, we examine how structured planning,
persistent state, and verification with evidence checks contribute to
this performance.
(3) \textbf{Evaluation beyond answer correctness.}
We propose observed support, an evaluation of whether each harness's correct
answers are recoverable from its saved evidence. We show that accounting
for this evidence changes how harnesses compare and further strengthens \paninidr{}'s advantage over existing
approaches. We believe these contributions offer a path toward more
reliable deep research systems.

\section{The \paninidr{} Harness}
\label{sec:paninidr}

The \paninidr{} harness coordinates research and verification through a shared
workspace that records the subquestions to resolve, their dependencies,
and the findings obtained so far. We use the Camp David example in
Figure~\ref{fig:paninidr-overview} to illustrate how this workspace evolves
during an investigation. The remainder of this section describes how
the workspace is constructed and used to guide research, verification,
and answer generation.

\subsection{Constructing the Research Workspace}
\label{sec:gsw}

Given a query $q$, the harness invokes an \emph{Orchestrator} as its main
agent. The Orchestrator oversees the creation and maintenance of research
state, launches sub-agents to investigate subquestions, and submits the
final answer. To initialize the workspace, it first calls a
\emph{Planner} agent to produce a structured research plan $P_0$ using
the \gsw{} representation introduced by
\citet{rajesh2026gsw,rajesh2026panini}. The plan records information
supplied by the query, the subquestions to resolve, and the relationships
and computations that establish dependencies between their answers.
Our choice of \gsw{}
is motivated by its situational representation of entities, their roles
and states, and their relationships. This structure provides explicit
context for verifying whether a proposed answer satisfies the conditions
of the subquestion it is intended to resolve.

The dependency structure encoded in $P_0$ forms a directed acyclic graph,
which the Orchestrator uses to guide research. Its nodes correspond to
the subquestions and a final answer target $v_\star$. An edge $u\rightarrow v$
indicates that the finding at $u$ supplies information used to resolve $v$.
For example, identifying the president provides the context for
investigating his wife, whose identity then guides the search for her
birth date. Nodes can be resolved through research or computation: in
Figure~\ref{fig:paninidr-overview}, the signing date and birth date supply
the inputs for computing the final answer, the wife's age.

The unknown answer at each node is represented as a \emph{blank} in the
\gsw{} plan. The Planner expresses the corresponding subquestion in
natural language as a \emph{retrieval hint}, incorporating the blank's
relations and relevant query context. For example, the relation ``was
born on'' between the wife node $u$ and birth-date blank $v$ gives the
dependency $u\rightarrow v$. The retrieval hint ``When was [wife] born?''
is attached to $v$, and the placeholder is replaced with her identity
once established. Each blank also carries metadata to support
verification, including an expected answer type, such as ``date,'' and
a verification claim describing what a proposed answer must satisfy,
such as ``the proposed date is the birth date of [wife].''

Before launching research, the harness checks the plan for structural
errors, including invalid references, missing or multiple answer targets,
and circular dependencies. It also checks whether retrieval hints are
consistent with the plan's dependencies. For example, the birth-date
hint may require the wife's identity without the plan recording that
dependency. A separate LLM call proposes corrections within a bounded
repair procedure, and the harness checks these corrections before
applying them. These plan checks are separate from the Verifier
(Section~\ref{sec:gate}), which reviews proposed findings rather than the plan.

\input{figures/cotinuum_overview.tex}

After checking the plan, the harness initializes the shared research
workspace by pairing the plan with per-node research records. These
records store findings, supporting evidence, and review status. We denote
the workspace at step $t$ by $W_t=(P_t,S_t)$, where $P_t$ is the current
plan and $S_t$ is the collection of research records linked to its nodes.
The initial workspace is $W_0=(P_0,S_0)$, with every blank marked as
unresolved in $S_0$; information supplied by the query is already
represented in $P_0$. The records are updated as research progresses.
For example, a record for the wife subquestion could contain the answer
\emph{Rosalynn Carter}, references to source passages establishing that
she was \emph{Jimmy Carter}'s wife, and a verified review status.
The Orchestrator uses these recorded findings to guide subsequent work.
Appendix~\ref{app:workspace-schema} gives the schema and scheduling rules.

\subsection{Researching a Subquestion}
\label{sec:harness}
\label{sec:researcher}

The Orchestrator uses the current workspace $W_t$ and the plan's
dependencies to select subquestions to investigate, launching a
\emph{Researcher} agent for each assignment.
Each Researcher receives the assigned subquestion, relevant plan
context, and available findings. In Figure~\ref{fig:paninidr-overview},
the signing-date and president subquestions can be investigated in
parallel.

Although the overall query may require multiple searches and reasoning
steps, an individual subquestion may be answerable from evidence obtained
in a single round of retrieval. The Researcher therefore begins with
an initial retrieval step, substituting available findings into the
retrieval hint and performing one search. For example, ``When was [wife]
born?'' becomes ``When was Rosalynn Carter born?'' once her identity is
established. The Researcher uses an LLM to propose a candidate answer
with supporting citations from the retrieved passages. If the candidate
passes type checking and verification (Section~\ref{sec:gate}), the
finding is recorded in the workspace. In
Figure~\ref{fig:paninidr-overview}, this initial step establishes the
signing date.

If the context required by the hint is unavailable, no candidate is
found, or the candidate fails these checks, the Researcher continues
with iterative search and reasoning using ReAct \citep{yao2023react},
retaining any candidate, sources, and feedback already obtained. It
chooses searches, reads sources, and submits proposals within its turn
allowance. To limit source text accumulating in the Researcher's context,
its reading tool uses a separate LLM agent to extract relevant information
and supporting quotations. The president subquestion in
Figure~\ref{fig:paninidr-overview} illustrates this extended research path.

An assignment may end before its subquestion is resolved. To support
later attempts, the harness retains candidates, retrieved sources,
previous searches, and verification feedback for that subquestion.
Relevant history is provided when the subquestion is assigned again,
allowing the Researcher to build on earlier work.
The Researcher submits candidate answers with supporting evidence for
verification. The resulting feedback can prompt further research before
a finding is recorded in the shared workspace.

\subsection{Verifying Findings and Updating the Workspace}
\label{sec:gate}

To verify a proposed answer, the harness combines deterministic checks
with LLM-based assessment. It first checks the expected answer type and
required evidence attachments. Establishing that the cited passages
support the answer requires further interpretation---for example,
whether a date refers to the birth of the intended person. A
\emph{Verifier} agent therefore evaluates the candidate against the
subquestion's verification claim, using the cited evidence and available
findings.

The Verifier checks whether the cited evidence supports the candidate as
an answer to the subquestion and whether it distinguishes that candidate
from plausible alternatives. Both judgments must be positive for the
Verifier to endorse the answer. For the president subquestion in
Figure~\ref{fig:paninidr-overview}, the evidence must establish who held
office on the signing date. The hypothetical proposal of Ronald Reagan
would fail because his dates in office do not match that date. When
verification fails, the Verifier returns feedback identifying missing
support or ambiguity. The Researcher can use this feedback to gather
further evidence or revise its proposal, with a limited number of retries
per assignment. Because each review concerns an individual subquestion,
proposals from different Researchers can be checked in parallel.

The harness admits a finding by filling the corresponding blank in the
workspace and recording its evidence and review status
(Appendix~\ref{app:execution-details}). In the
$W_0\!\rightarrow W_1$ transition in
Figure~\ref{fig:paninidr-overview}, admitting the signing date and Jimmy
Carter's identity provides a date for the final calculation and the
context needed to investigate his wife.

\subsection{Answering the Query}
\label{sec:completion}

As findings are admitted, they guide subsequent research and provide
inputs for computation. In Figure~\ref{fig:paninidr-overview}, identifying
Rosalynn Carter at $W_2$ supplies the context needed to investigate her
birth date. The harness executes each planned computation in code once
its inputs are available. Here, the birth date of 18 August 1927 and
signing date of 17 September 1978 yield 51 completed years: her birthday
has already occurred in the signing year. The age fills the answer target
at $W_3$. Once the harness checks that the target is resolved, the
Orchestrator submits the answer.

The investigation in Figure~\ref{fig:paninidr-overview} completes its
original plan, but a Researcher may discover that answering its assigned
subquestion requires an intermediate fact missing from the plan. It can
then end its assignment and return a suggested additional subquestion,
together with an explanation of why it is needed, to the Orchestrator.
The Orchestrator can ask the Planner to revise the existing plan using
the findings and the Researcher's suggestion. Revisions can add or change
subquestions and reopen affected findings for further research, while
retaining the remaining records. Research then continues on the revised
plan. If the overall investigation ends without a final answer to the
original query, for example after exhausting its research allowance, the
harness can make a final LLM call to synthesize that answer from the
collected evidence, without further search
(Appendix~\ref{app:execution-details}).

\begingroup
\renewcommand{\paninidr}{\textsc{Panini-DR}}
\section{Experimental Setup}
\label{sec:setup}

\subsection{Datasets}
\label{sec:datasets}

We use three complementary question-answering benchmarks.
BrowseComp-Plus (BCP) tests identifying an entity from detailed clues
\citep{chen2025browsecompplus}; FRAMES tests combining facts across
sources \citep{krishna2024frames}; and MoNaCo includes collection-valued
questions, where recovering the required items without adding incorrect
ones is important \citep{wolfson2025monaco}.
For BCP, we use the official 100K-document corpus and the Context-Folding
CF-150 split, with 50 questions in each of the easy, medium, and hard
groups \citep{sun2025contextfolding}. We evaluate on all 824 FRAMES
questions and all 1,315 MoNaCo questions.

\subsection{Evaluation Metrics}
\label{sec:eval-metrics}

\noindent\textbf{Answer quality.}
We report \emph{official benchmark scores}: accuracy on BCP and FRAMES
and mean item-level F1 on MoNaCo. All systems receive 3
\texttt{gpt-5.6-luna} judgments of each answer, aggregated by majority
vote on BCP and FRAMES and maximum F1 over complete assessments on
MoNaCo, since item matching can spuriously lower F1 in different assessments. Every question remains in the
denominator; missing judgments score 0. Appendix~\ref{app:protocol}
details scoring and a second-grader sensitivity check.

\noindent\textbf{Observed support.}
Accuracy alone does not show whether the agent has evidence for its answer
\citep{rashkin2023attribution,gao2023alce}.
We propose \emph{observed support} (Obs.), which keeps answer credit only when the credited
content is recoverable from the system's saved evidence. For BCP and FRAMES,
let $c_i$ indicate that answer $i$ is officially correct and $s_i$
indicate that every checkable answer part is recoverable from its
saved evidence. Over $N$ questions,
\begin{equation}
\operatorname{Obs}=\frac{1}{N}\sum_{i=1}^{N}c_i s_i.
\label{eq:observed-supported}
\end{equation}
For MoNaCo, we recompute F1 using only judge-credited items that pass
the support rule. 
Obs.\ measures recoverability through deterministic matching, and
Appendix~\ref{app:derivation-audit} extends it to computed answers.
Section~\ref{sec:observed-results} separately tests whether the evidence
supports the requested claim.
Appendix~\ref{app:observed-support} details the support scorer.

\noindent\textbf{Research cost.}
Locally served models reuse cached prefix states, so submitted-token totals
overstate computation. We therefore also estimate computation from
uncached input and generated tokens
(Section~\ref{sec:cost}).

\subsection{Baselines}
\label{sec:baselines}

We compare against seven harnesses with different ways of organizing
research and retaining findings. \textbf{PyRAG} \citep{sun2026pyrag}
executes generated retrieval programs; \textbf{Re-TRAC}
\citep{zhu2026retrac} compresses research trajectories; and
\textbf{IterResearch} \citep{chen2025iterresearch} maintains an evolving
research report. \textbf{SLEUTH} \citep{liu2026sleuth} tracks facts,
ranked hypotheses, and open questions; \textbf{Omni-Decision}
\citep{ma2026omnidecision} maintains query-specific evidence needs;
and \textbf{HyMem} \citep{wang2026hymem} manages episode, working,
and tool memories through folding and selective injection into later
calls. \textbf{Open Deep Search}
\citep{alzubi2025opendeepsearch} provides a CodeAct search-and-tool agent.

We use released implementations where available. For the prompt-based
SLEUTH, we follow the authors' published prompts and reasoning protocol;
we reimplement Omni-Decision and HyMem from their published specifications.
Appendix~\ref{app:compat} documents adaptations to the shared environment.

\subsection{Implementation Details}
\label{sec:implementation}

All systems use \texttt{gpt-oss-120b} \citep{openai2025gptoss} and, within
each benchmark, the same frozen corpus and retrieval service: BM25 and
dense \texttt{Qwen3-Embedding-8B} retrieval, followed by
\texttt{jina-reranker-v3.5}. Baselines access it through a shared search
gateway that returns full documents on BCP and chunks on FRAMES and MoNaCo.
SLEUTH keeps its published top-3 setting, chosen to avoid context dilution
\citep{liu2026sleuth}; other baselines use published retrieval breadth where
specified and 6 hits otherwise, and \paninidr{} searches return 6 reranked
snippets. Baselines keep their published or released-code research
allowances; \paninidr{} allows each Researcher assignment 20 ReAct turns and up to 5 retries after
failed checks (Appendix~\ref{app:protocol}).

\section{Results and Discussion}
\label{sec:results}

We assess answer quality and retained evidence, then test the contributions
of research structure and verification through ablations. Cost and
allowance studies examine the resources these gains need.

\subsection{Answer Quality}
\label{sec:answer-quality}

\input{tables/answer_quality.tex}

\paninidr{} has the best average across the three benchmarks, and its
advantage is largest on the hardest questions (Table~\ref{tab:main-results}).
Because all systems share the backbone and retrieval service, these
differences come from the harnesses. On the BCP hard group, which ReAct
agents solve least often, \paninidr{} nearly doubles SLEUTH's accuracy. The
planning ablation points the same way, since adding the plan and persistent
state doubles the hard questions answered
(Appendix~\ref{app:ablation-workspace}). The 16-point BCP gain over SLEUTH
holds when we remove the questions SLEUTH refuses for input length and is
consistent across three runs per harness (Appendix~\ref{app:uncertainty}).
On MoNaCo, HyMem has the highest F1 and \paninidr{} ranks fifth, although our collection answers are
the most precise. Once answers must be backed by saved evidence,
\paninidr{} also leads on MoNaCo (Section~\ref{sec:observed-results}).

\subsection{Evidence Quality}
\label{sec:observed-results}

\input{tables/observed_support.tex}

Recording the evidence behind each finding should leave answers that readers can check against their sources. When correct answer content must be recoverable from saved evidence,
\paninidr{} leads Obs.\ on all three benchmarks, and its margin over SLEUTH, the
strongest baseline, widens compared with official scores
(Table~\ref{tab:main-observed-support}).
A model-based claim-support audit tests whether the saved evidence supports
the requested claims. It rebuilds each answer from the supported claims
only. \paninidr{} remains ahead of SLEUTH on all three benchmarks under this
audit (Appendix~\ref{app:grounded-only-audit}).
We also examine how research finds and retains relevant sources.
On BCP, \paninidr{} reaches most of the annotated source pages, while SLEUTH
reaches fewer than half, and more of the pages it cites are annotated. Both gains also hold on FRAMES and MoNaCo
(Appendices~\ref{app:evidence-coverage} and~\ref{app:gold-reference}).
A matched trajectory shows how findings guide further research.
\paninidr{} identifies Jim Bridger from biographical clues and then uses the
name to find the expedition Bridger joined, while SLEUTH never establishes
the identity.
Finally, we test whether the retained evidence is useful beyond our own
answer generation. In a common-reader test, the same \texttt{gpt-oss-120b}
reader answers from the evidence of four systems. Our evidence yields the highest
answer quality on all three benchmarks, whether the reader is told an answer is present or may abstain
(Appendix~\ref{app:fixed-reader-framing}).

\subsection{Ablations}
\label{sec:ablations}

We first test whether structured planning and persistent research state
improve answers beyond prose notes. Both reduced harnesses retain Researchers and the Verifier: \textsc{Off} keeps notes and Researcher
reports, while \textsc{On} adds a plan, findings linked to its subquestions, and plan
revision (Appendix~\ref{app:plan-revision-analysis}). Under the same configured limits, \textsc{On} raises
BCP accuracy from 41.3\% to 58.0\%. The 16.7-point gain persists when both use the same final answer synthesis (56.7\% versus 73.3\%).
To check whether findings guide later work, we track newly recorded entities absent from the question and followed by another research
assignment. With planning and state, 77.4\% appear in a later search,
compared with 32.1\% without them: earlier findings more often become
inputs to subsequent research
(Appendix~\ref{app:ablation-workspace}).

Because errors in these findings can propagate, we next test verification
before reuse. With the workspace and research tools fixed, the full verification policy with up to 5 retries adds 16.0 points to BCP accuracy and Obs. compared with removing the Verifier and the citation and evidence-attachment checks. Assignment and answer-type checks remain in both conditions. Gains also hold
on the matched FRAMES and MoNaCo subsets
(Appendix~\ref{app:ablation-verification}).
The proposal histories show how feedback changes the investigation: 77 of 101 proposals endorsed after an earlier rejection change the
answer or cited sources. For example, rejecting an editor's surname
leads to a full name supported by credits from both requested films
(Appendix~\ref{app:ablation-cap}). Verification thus helps improve
findings through further evidence gathering before they guide later work.

\subsection{Computation and Research Allowances}
\label{sec:cost}

\paninidr{} sends more prompt tokens than most baselines, though fewer than
Re-TRAC (Table~\ref{tab:cost} in Appendix~\ref{app:cost-profile}). Most of
these tokens go to the Researcher's ReAct search and its reading tool, and
many calls resend context from earlier calls. The vLLM server caches
recently processed prompt prefixes, so it does not recompute resent context.
In separate BCP runs where we recorded its cache counters, 66.5\% of our
input tokens came from this cache. Our estimated computation is therefore
$7.3\times$ SLEUTH's, although we send $19.2\times$ its prompt tokens
(Appendix~\ref{app:compute-cost}).

We use this estimate to test whether repeated research with similar
computation can close the accuracy gap. We run SLEUTH 8 times per question
and take its most common non-empty answer. This uses $1.12\times$ our
computation but reaches 64.7\% BCP accuracy, versus 73.3\% for \paninidr{}
(Appendix~\ref{app:matched-compute}). Instead of more independent runs, we
also give each investigation more turns to build on its findings. Raising
the Researcher's turn allowance from 20 to 50 and 80 turns improves BCP
accuracy from 73.3\% to 74.7\% and 76.0\%, while SLEUTH shows no steady
gain (Appendix~\ref{app:ablation-budget}). However, 80 turns uses
$2.8\times$ the default's prompt tokens for 2.7 more points, so we keep 20 turns.

\FloatBarrier
\endgroup

\section{Related Work}
\label{sec:related}

\noindent\textbf{Research state and executable plans.}
Chain-of-thought prompting makes intermediate reasoning explicit within a
model call \citep{wei2022chainofthought}.
Self-Ask and IRCoT connect question decomposition to retrieval
\citep{press2023selfask,trivedi2023ircot}. ReAct interleaves reasoning and tool
use \citep{yao2023react}, while ReWOO
and LLMCompiler make tool dependencies explicit in execution plans
\citep{xu2023rewoo,kim2024llmcompiler}. Several of our baselines preserve progress in evolving reports, compressed
trajectories, or memories \citep{chen2025iterresearch,zhu2026retrac,wang2026hymem};
SLEUTH and Omni-Decision also maintain explicit evidence state with checked
updates \citep{liu2026sleuth,ma2026omnidecision}. \paninidr{} builds on the
\gsw{} representation \citep{rajesh2026gsw,rajesh2026panini}; its
contribution is the execution protocol connecting typed research
assignments, subquestion-scoped review, and admitted values used by
subsequent work, rather than a dependency graph alone.

\noindent\textbf{Verification during research.}
Process supervision and verifier-guided search use intermediate checking to
improve reasoning \citep{lightman2023verifystep,snell2024scaling}.
Chain-of-Verification revises drafts using separate checking questions
\citep{dhuliawala2024cove}.
Self-RAG learns retrieval and generation critiques \citep{asai2024selfrag};
CRAG evaluates retrieved material to guide corrective retrieval
\citep{yan2024crag}. LLM-as-a-Verifier studies fine-grained, decomposed
model-based evaluation \citep{kwok2026llmverifier}.
\paninidr{} connects this checking role to persistent research subquestions:
the Verifier evaluates candidates against their subquestions' verification
claims and cited evidence, and the harness determines admission while
retaining the review status.

\noindent\textbf{Attribution and support.}
Attribution work evaluates whether sources support generated claims
\citep{rashkin2023attribution,gao2023alce,malaviya2024expertqa}.
Observed support complements this with deterministic answer
recoverability from saved evidence. Appendix~\ref{app:related-work} extends
this discussion.

\section{Limitations and Concluding Remarks}
\label{sec:limitations}

In this work, we introduce \paninidr{}, a multi-agent harness for deep
research built around two ideas: maintaining explicit research state and
verifying intermediate findings before they guide subsequent work. We
implement these ideas using the \gsw{} representation to organize
subquestions, their dependencies, findings, and supporting evidence in a
shared workspace. The workspace allows the Orchestrator to coordinate
the investigation while separate agents research and verify individual
subquestions. Across three deep research benchmarks, \paninidr{} achieves
the strongest average answer quality, with particularly large gains on
harder BrowseComp-Plus questions. Ablations support the value of both
structured research state and intermediate verification. Our
observed-support evaluation further shows that this advantage widens when
correct answer content must be recoverable from the retained evidence.

However, we identify several limitations to our work. Verification
depends on the subquestions identified by the Planner, the candidates
considered during research, and the evidence available to the Verifier.
Model-based judgments can be incorrect, and checking individual findings
cannot guarantee that the plan covers everything needed to answer the
original query. Our primary comparisons use one research backbone and
fixed benchmark corpora; generalization to other models and open-web
research remains untested. Observed support measures answer recoverability
from saved evidence, whose coverage and format vary across systems
(Appendix~\ref{app:observed-support}). It does not establish that every
claim is supported, and our complementary claim-support audits also rely
on model judgments. The harness requires substantial computation, while
increasing the verification allowance does not consistently improve answer
quality (Section~\ref{sec:cost}; Appendix~\ref{app:ablation-cap}). In future
work, we plan to expand our study to a wider range of models and open-web
datasets. Developing reliable, computationally efficient methods for
scaling verification is therefore an important direction for future
research.

\clearpage
\section*{Reproducibility Statement}
\label{sec:repro}

Appendix~\ref{app:protocol} records the frozen corpora, shared backbone,
retrieval service, run limits, and judging protocol for every comparison;
Appendix~\ref{app:compat} records all baseline compatibility changes. The
observed-support scorer deterministically combines each system's saved traces,
source references, and cached judgments, as specified in
Appendix~\ref{app:observed-support}. The release will contain the harness,
gateway, scorer, evaluation configurations, and analysis scripts in a
public GitHub repository after the anonymous review process.

\section*{Ethics Statement}

This work aims to improve the reliability of deep research systems and evaluates the proposed methods on existing benchmark datasets. We do not identify additional ethical concerns specific to this study beyond those associated with language models and their applications.

\section*{AI Use Statement}

Language models were used as components of the systems studied and as
graders, as described in Section~\ref{sec:setup}. The authors also used
AI assistance to edit the draft and to catch grammatical errors. The authors take full responsibility for the contents of the paper.

\bibliographystyle{iclr2027_conference}
\label{sec:refs}
\bibliography{references}

\clearpage
\appendix

\section*{Appendix Contents}

\noindent
\begin{tabularx}{\linewidth}{@{}lXr@{}}
\toprule
\textbf{Section} & \textbf{Contents} & \textbf{Page} \\
\midrule
\ref{app:protocol} &
\hyperref[app:protocol]{Evaluation Protocol Details} &
\pageref{app:protocol} \\
\quad\ref{app:shared-setup} &
\hyperref[app:shared-setup]{Shared Evaluation Setup} &
\pageref{app:shared-setup} \\
\quad\ref{app:answer-grading} &
\hyperref[app:answer-grading]{Answer Generation and Grading} &
\pageref{app:answer-grading} \\
\quad\ref{app:grader-sensitivity} &
\hyperref[app:grader-sensitivity]{Sensitivity to the Grading Model} &
\pageref{app:grader-sensitivity} \\
\ref{app:implementation} &
\hyperref[app:implementation]{Implementation Through Recorded Investigations} &
\pageref{app:implementation} \\
\quad\ref{app:workspace-schema} &
\hyperref[app:workspace-schema]{From the Question to a Research Plan} &
\pageref{app:workspace-schema} \\
\quad\ref{app:brazuca-example} &
\hyperref[app:brazuca-example]{Brazuca: Findings, Feedback, and Evidence Completion} &
\pageref{app:brazuca-example} \\
\quad\ref{app:other-execution-paths} &
\hyperref[app:other-execution-paths]{When Research Takes a Different Path} &
\pageref{app:other-execution-paths} \\
\quad\ref{app:example-record} &
\hyperref[app:example-record]{Fort Santiago: Answering from the Retained Research} &
\pageref{app:example-record} \\
\ref{app:observed-support} &
\hyperref[app:observed-support]{Evidence Quality and Validation} &
\pageref{app:observed-support} \\
\quad\ref{app:scorer-sensitivity} &
\hyperref[app:scorer-sensitivity]{Checking the Scorer and Evidence Record} &
\pageref{app:scorer-sensitivity} \\
\quad\ref{app:grounded-only-audit} &
\hyperref[app:grounded-only-audit]{Model-Based Grounded-Only Audit} &
\pageref{app:grounded-only-audit} \\
\quad\ref{app:fixed-reader-framing} &
\hyperref[app:fixed-reader-framing]{Answering from Each System's Evidence} &
\pageref{app:fixed-reader-framing} \\
\quad\ref{app:derivation-audit} &
\hyperref[app:derivation-audit]{Checking Computed Answers} &
\pageref{app:derivation-audit} \\
\ref{app:additional} &
\hyperref[app:additional]{Additional Experimental Results} &
\pageref{app:additional} \\
\quad\ref{app:diagnostics} &
\hyperref[app:diagnostics]{Research Evidence and Answer Completion} &
\pageref{app:diagnostics} \\
\qquad\ref{app:evidence-coverage} &
\hyperref[app:evidence-coverage]{Finding Sources and Connecting Findings} &
\pageref{app:evidence-coverage} \\
\qquad\ref{app:recovery-analysis} &
\hyperref[app:recovery-analysis]{Revising the Plan and Recovering an Answer} &
\pageref{app:recovery-analysis} \\
\qquad\ref{app:gold-reference} &
\hyperref[app:gold-reference]{Answering with Annotated Sources} &
\pageref{app:gold-reference} \\
\quad\ref{app:ablations} &
\hyperref[app:ablations]{Planning, Verification, and Recovery Controls} &
\pageref{app:ablations} \\
\qquad\ref{app:ablation-workspace} &
\hyperref[app:ablation-workspace]{Structured Planning and Persistent State} &
\pageref{app:ablation-workspace} \\
\qquad\ref{app:ablation-verification} &
\hyperref[app:ablation-verification]{Verification Before Reuse} &
\pageref{app:ablation-verification} \\
\qquad\ref{app:ablation-cap} &
\hyperref[app:ablation-cap]{Responding to Verification Feedback} &
\pageref{app:ablation-cap} \\
\quad\ref{app:computation} &
\hyperref[app:computation]{Computation and Research Allowances} &
\pageref{app:computation} \\
\qquad\ref{app:cost-profile} &
\hyperref[app:cost-profile]{Recorded Tokens and Cache Reuse} &
\pageref{app:cost-profile} \\
\qquad\ref{app:matched-compute} &
\hyperref[app:matched-compute]{Repeated Sampling at Comparable Computation} &
\pageref{app:matched-compute} \\
\qquad\ref{app:ablation-budget} &
\hyperref[app:ablation-budget]{Extending Individual Investigations} &
\pageref{app:ablation-budget} \\
\quad\ref{app:uncertainty} &
\hyperref[app:uncertainty]{Run Stability, Paired Comparisons, and Answer Structure} &
\pageref{app:uncertainty} \\
\quad\ref{app:compat} &
\hyperref[app:compat]{Baseline Adaptations} &
\pageref{app:compat} \\
\ref{app:related-work} &
\hyperref[app:related-work]{Extended Related Work} &
\pageref{app:related-work} \\
\quad\ref{app:related-plans} &
\hyperref[app:related-plans]{Reasoning Structure, Plans, and Semantic Workspaces} &
\pageref{app:related-plans} \\
\quad\ref{app:related-harnesses} &
\hyperref[app:related-harnesses]{Deep-Research Harnesses and Persistent Memory} &
\pageref{app:related-harnesses} \\
\quad\ref{app:related-verification} &
\hyperref[app:related-verification]{Verification, Feedback, and Corrective Retrieval} &
\pageref{app:related-verification} \\
\quad\ref{app:related-attribution} &
\hyperref[app:related-attribution]{Attribution and Evidence-Based Evaluation} &
\pageref{app:related-attribution} \\
\bottomrule
\end{tabularx}

\clearpage
\setcounter{figure}{0}
\renewcommand{\thefigure}{A\arabic{figure}}
\renewcommand{\theHfigure}{A.\arabic{figure}}

\section{Evaluation Protocol Details}
\label{app:protocol}

\subsection{Shared Evaluation Setup}
\label{app:shared-setup}

To compare complete research harnesses on a common foundation, we share
the backbone, fixed corpus, and retrieval service within each benchmark.
The CF-150 difficulty groups in Section~\ref{sec:datasets} are based on
ReAct accuracy over 8 runs. FRAMES questions draw on 2 to 15 articles;
Table~\ref{tab:monaco-decomposition} separates MoNaCo's single-item and
collection questions.

The shared backbone is \texttt{gpt-oss-120b}, served by vLLM 0.24 with
MXFP4 weights, temperature 0, seed 42, and medium reasoning effort unless
a system sets its own. All systems use BM25 and Qwen3-Embedding-8B retrieval
followed by \texttt{jina-reranker-v3.5}. The baseline search gateway returns
full documents on BCP and chunks on FRAMES and MoNaCo; on BCP, this adapts
SLEUTH's original paragraph retrieval \citep{liu2026sleuth}.
\paninidr{} returns 6 snippets per search, with source reading
handled separately (Appendix~\ref{app:execution-details}).

Across all 3 benchmarks, \paninidr{} uses medium effort for the Planner,
Researcher's ReAct stage, Reader, and Verifier, with 20 ReAct turns per Researcher
assignment, 12 Orchestrator turns, and 5 plan-update requests. Its shared
review cap of 5 limits failed-check re-attempts, not total
Researcher turns or Verifier calls.
Baselines retain their own budgets (Table~\ref{tab:baseline-protocols}):
Re-TRAC, Omni-Decision, SLEUTH, and HyMem follow paper-reported values;
IterResearch and Open Deep Search use released-code defaults. Retrieval
breadth follows published values where specified and otherwise defaults
to 6. SLEUTH retains its published top-3 setting to avoid diluting useful
evidence with additional retrieved text \citep{liu2026sleuth}.
These comparisons do not equalize computation; Appendix~\ref{app:matched-compute}
tests additional sampling at comparable computation.
Appendix~\ref{app:compat} documents code adaptations and paper-based
reimplementations.

\input{tables/baseline_protocols.tex}

\subsection{Answer Generation and Grading}
\label{app:answer-grading}

The headline comparisons use 1 research run per configuration, producing
1 answer per question. We grade that same answer 3 times with
\texttt{gpt-5.6-luna}, replacing the benchmark judge model while retaining
the BrowseComp-Plus criteria for BCP and FRAMES and the official MoNaCo
criteria. BCP and FRAMES use majority voting over the 3 judgments of
full-answer correctness.

For MoNaCo, a grading pass can overlook valid item matches. We therefore
select the highest-F1 complete judgment of the 3 passes, identically for
every system, and use its item-level credits for both official F1 and
support accounting. Every question remains in the denominator;
$N$ counts questions, not grading passes.\footnote{Blank FRAMES answers
score 0 without a grader call. Missing judgments score 0; unreadable
judgments do not count as correct. Otherwise, we keep the official
judgment, even when it gives credit to an empty answer or an abstention.}

The sampling control uses these same grading rules to test whether repeated
SLEUTH runs can match \paninidr{} at comparable computation
(Appendix~\ref{app:matched-compute}, Table~\ref{tab:matched-compute}).
For each question, we select the most frequent non-empty answer across
runs. This answer-selection vote is separate from the 3 grading judgments;
accuracy and Obs.\ use the selected run's verdict and saved evidence.

\subsection{Sensitivity to the Grading Model}
\label{app:grader-sensitivity}

Because answer correctness is assessed by an LLM judge, we test whether
\paninidr{}'s BCP lead depends on the cost-focused \texttt{gpt-5.6-luna}
grader. We re-grade the same saved answers for all 150 questions from
5 systems with \texttt{gpt-5.6-terra}, which is designed to balance
capability and cost,\footnote{OpenAI's model descriptions:
\href{https://developers.openai.com/api/docs/models/gpt-5.6-luna}{GPT-5.6 Luna}
and \href{https://developers.openai.com/api/docs/models/gpt-5.6-terra}{GPT-5.6 Terra}.}
keeping all other grading settings fixed
(Table~\ref{tab:judge-sensitivity}). Four systems
receive identical verdicts on every question; \paninidr{} gains
1 correct answer. The BCP ranking therefore survives the grader change,
with the lead over the strongest baseline increasing from 24 to 25 questions.

\input{tables/judge_sensitivity.tex}

\section{Implementation Through Recorded Investigations}
\label{app:implementation}
\label{app:execution-details}

\input{appendices/implementation_trace.tex}

\section{Evidence Quality and Validation}
\label{app:observed-support}

\input{appendices/evidence_scoring.tex}
\input{appendices/scorer_sensitivity.tex}
\input{appendices/grounded_only_audit.tex}

\input{appendices/fixed_reader_framing.tex}

\input{appendices/obs_d.tex}

\section{Additional Experimental Results}
\label{app:additional}

\input{appendices/additional_results.tex}

\FloatBarrier
\input{appendices/related_work.tex}

\end{document}

%% file: math_commands.tex
\usepackage{amsmath,amsfonts,bm}

\def\eqref#1{equation~\ref{#1}}

\def\1{\bm{1}}

\DeclareMathAlphabet{\mathsfit}{\encodingdefault}{\sfdefault}{m}{sl}
\SetMathAlphabet{\mathsfit}{bold}{\encodingdefault}{\sfdefault}{bx}{n}

%% file: figures/cotinuum_overview.tex
\begin{figure}[htbp]
  \centering
  \includegraphics[width=\linewidth]{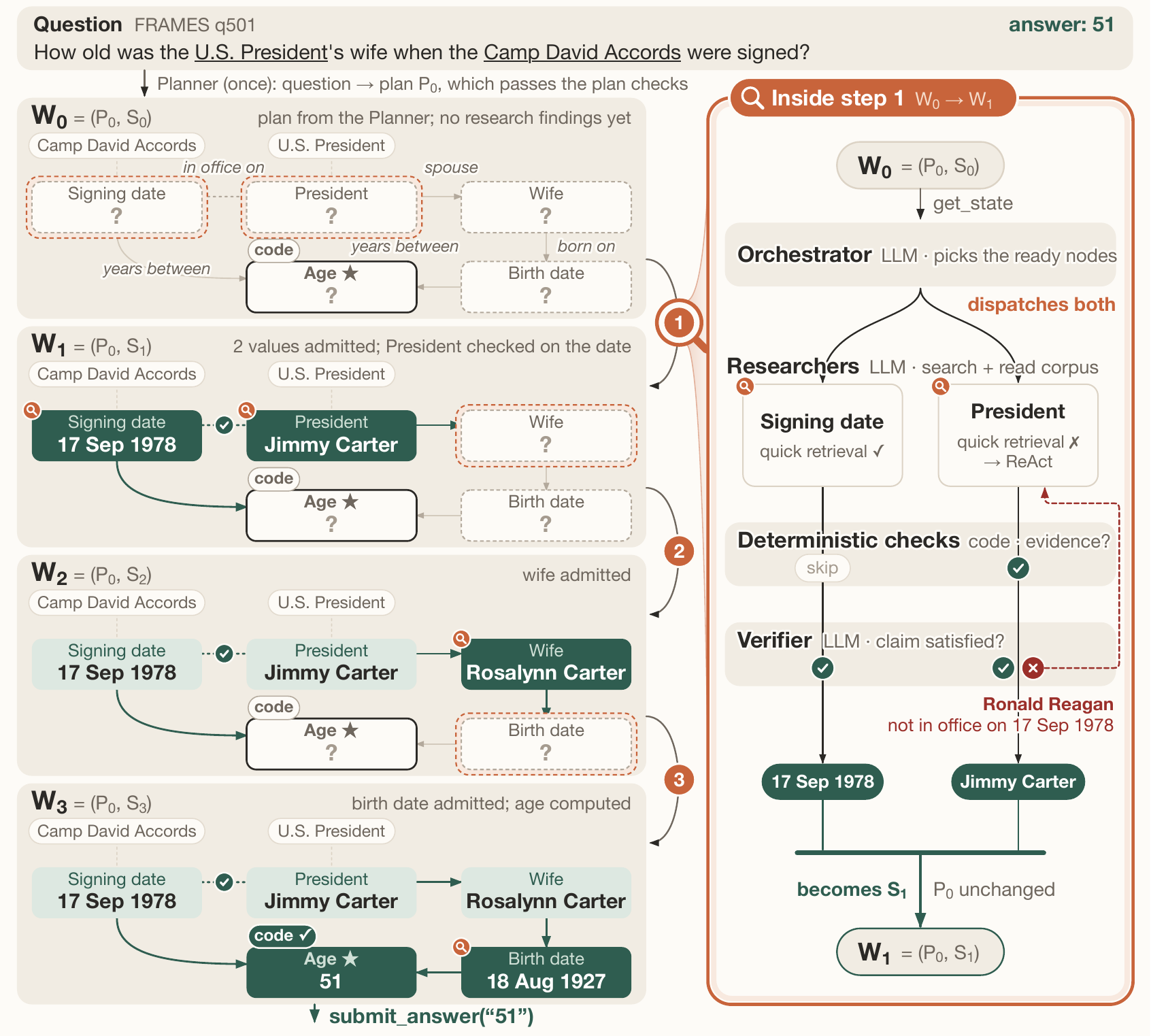}
  \caption{\paninidr{} illustrated on a sample query.
  The schematic stages separate the subquestions for explanation, rather
  than reproduce the recorded dispatch timing. Each $W_k=(P,S_k)$ pairs
  the plan with per-node research records; $\varnothing$ denotes that no
  findings have yet been recorded. Dark green marks additions, dashed
  boxes remain unresolved, and orange outlines indicate readiness.
  The expanded stage shows each Researcher's initial retrieval step
  (labelled ``typed resolver'') and, when needed, ReAct agent, followed
  by verification and admission; the Reagan rejection is hypothetical. Findings supply
  subsequent subquestions and the date calculation that produces 51.
  The recorded execution investigated all four factual subquestions in
  one dispatch.}
  \label{fig:paninidr-overview}
\end{figure}

%% file: tables/answer_quality.tex
\begin{table}[!htbp]
\centering
\small
\setlength{\tabcolsep}{4pt}
\begin{tabular}{lrrrrrrr}
\toprule
& \multicolumn{4}{c}{\textbf{BCP CF-150}} & \textbf{FRAMES} & \textbf{MoNaCo} & \textbf{Avg.} \\
\cmidrule(lr){2-5}\cmidrule(lr){6-6}\cmidrule(lr){7-7}\cmidrule(lr){8-8}
\textbf{Method} & Easy & Med. & Hard & Overall & Accuracy & F1 & \\
\midrule
PyRAG & 56 & 20 & 6 & 27.3 & 61.9 & 0.446 & 44.6 \\
Re-TRAC & 70 & 32 & 16 & 39.3 & 64.9 & 0.544 & 52.9 \\
Omni-Decision & 84 & 48 & 10 & 47.3 & 64.6 & 0.359 & 49.3 \\
HyMem & 82 & 52 & 12 & 48.7 & 70.8 & \textbf{0.612} & 60.2 \\
Open Deep Search & 82 & 50 & \underline{28} & 53.3 & 76.0 & \underline{0.571} & 62.1 \\
IterResearch & 84 & \underline{60} & 20 & 54.7 & 70.5 & 0.564 & 60.5 \\
SLEUTH & \underline{86} & 58 & \underline{28} & \underline{57.3} & \underline{77.2} & 0.556 & \underline{63.4} \\
\midrule
\textbf{\paninidr{}} & \textbf{92} & \textbf{76} & \textbf{52} & \textbf{73.3} & \textbf{79.1} & 0.549 & \textbf{69.1} \\
\bottomrule
\end{tabular}
\caption{Official accuracy (\%) on BCP and FRAMES and mean F1 on MoNaCo.
Avg. equally weights BCP Overall, FRAMES, and MoNaCo, with F1 rescaled
to 0--100. BCP groups contain 50 questions each. Bold: best;
underline: second best, including ties.}
\label{tab:main-results}
\end{table}

%% file: tables/observed_support.tex
\begin{table}[!htbp]
\centering
\small
\setlength{\tabcolsep}{11pt}
\begin{tabular}{lrrrr}
\toprule
\textbf{Method} & \textbf{BCP} & \textbf{FRAMES} & \textbf{MoNaCo} & \textbf{Avg.} \\
\midrule
PyRAG & 26.0 & 46.6 & 0.332 & 35.3 \\
Re-TRAC & 14.0 & 46.1 & \underline{0.396} & 33.2 \\
Omni-Decision & 45.3 & 53.9 & 0.275 & 42.2 \\
HyMem & 46.7 & 36.7 & 0.388 & 40.7 \\
Open Deep Search & 51.3 & 47.6 & 0.325 & 43.8 \\
IterResearch & 51.3 & 45.9 & 0.390 & 45.4 \\
SLEUTH & \underline{54.0} & \underline{55.2} & \underline{0.396} & \underline{49.6} \\
\midrule
\textbf{\paninidr{}} & \textbf{70.7} & \textbf{66.1} & \textbf{0.458} & \textbf{60.9} \\
\bottomrule
\end{tabular}
\caption{Observed support (Obs.): official answer credit recoverable
from saved evidence, as defined in Section~\ref{sec:eval-metrics}. BCP and
FRAMES report percentages; MoNaCo reports mean F1. Avg. equally weights
the three benchmarks, with F1 rescaled to 0--100. Best scores are bold;
second-best scores are underlined, including ties.}
\label{tab:main-observed-support}
\end{table}

%% file: tables/baseline_protocols.tex
\begin{table}[ht]
\centering
\small
\begin{tabular}{lp{9.8cm}}
\toprule
\textbf{System} & \textbf{Protocol} \\
\midrule
PyRAG & 1 executable retrieval program; no read tool. \\
Re-TRAC & 8 recursive search cycles, each compressed into the next cycle's starting state. \\
IterResearch & 25 turns; sampling temperature is hard-coded at 0.6. \\
Omni-Decision & 15 iterations behind an evidence-state readiness gate. \\
Open Deep Search & 15 CodeAct steps; search returns model-written summaries. \\
SLEUTH & Published default of 20 turns and retrieval top-$k=3$; maintains ranked hypotheses. \\
HyMem & 12 planner turns and 10 executor turns per task, with top-$k=6$. \\
\bottomrule
\end{tabular}
\caption{Baseline protocols used for the reported runs.}
\label{tab:baseline-protocols}
\end{table}

%% file: tables/judge_sensitivity.tex
\begin{table}[!htbp]
\centering
\small
\setlength{\tabcolsep}{7pt}
\begin{tabular}{lrrrrr}
\toprule
System & Primary grader & Second grader & $\Delta$ & Agreement & $\kappa$ \\
\midrule
\textbf{\paninidr{}} (cap 5) & 110 (73.3) & 111 (74.0) & $+0.7$ & 99.3 & 0.983 \\
SLEUTH & 86 (57.3) & 86 (57.3) & $0.0$ & 100 & 1.000 \\
HyMem & 73 (48.7) & 73 (48.7) & $0.0$ & 100 & 1.000 \\
Omni-Decision & 71 (47.3) & 71 (47.3) & $0.0$ & 100 & 1.000 \\
Re-TRAC & 59 (39.3) & 59 (39.3) & $0.0$ & 100 & 1.000 \\
\bottomrule
\end{tabular}
\caption{Grader sensitivity on BCP CF-150 (150 questions):
\texttt{gpt-5.6-luna} (primary) versus \texttt{gpt-5.6-terra}, with the
same answers, official prompt, and 3-pass majority. Scores are correct-answer
counts with accuracy (\%) in parentheses; $\Delta$ is in percentage points.
Agreement is the percentage of matching verdicts; $\kappa$ is Cohen's kappa.}
\label{tab:judge-sensitivity}
\end{table}

%% file: appendices/implementation_trace.tex
BCP q628 follows an investigation from its initial plan to the answer
\emph{Brazuca}, showing how Researchers obtain intermediate findings,
respond to feedback, and use earlier findings in later assignments.
We then examine collection answers, retries, plan revision, and recovery
through other recorded investigations.

\subsection{From the Question to a Research Plan}
\label{app:workspace-schema}

The full question is:
\begin{quote}
There is a short film, produced in 2017, with a runtime of 19 minutes,
that revolves around a young boy and his pursuit to win a special ball
and rise up in the eyes of his friends. The director of this film was
born before 1990 and after 1980, he studied Public Administration, and
this film is his third. The film's editor was also an editor for a 2016
film about murder solving. The production designer for the film in
question was also the art director for the same film.
\end{quote}
Finding a plausible title is therefore only part of the investigation: the research must
connect that title to the people, dates, and credits in the question.

\noindent\textbf{Making the clues researchable.}
The Planner connects the film to its year, runtime, director, editor,
and designer, then connects those people to the required biography and
credits. In this cap-5 execution, the plan has 9 given anchors, 13 relations,
and 12 plan nodes with unknown values (\emph{blanks}): the answer target, 7 intermediate
facts, and 4 Boolean outputs. Four deterministic constraints check runtime
equality, third-film equality, and the 2 birth-year bounds. Code evaluates
these comparisons once their inputs are available; they need no research
assignments.

Omitting step subscripts, write $P=(\mathcal{A},V,E,\mathcal{C},v_\star)$:
anchors, nodes, dependency edges, computations, and answer target.
$G=(V,E)$ is Section~\ref{sec:gsw}'s dependency graph. Each node
specifies its role, value type, relations, retrieval hint, and verification
claim; each computation names its function, inputs, and output. Before
research, the harness requires a nonempty plan, one target, valid references,
and no dependency cycle; LLM-proposed repairs must pass the same checks.
Advisory tags flag possible omissions or ambiguity without blocking dispatch
or changing the plan. These checks do not establish that the plan can
produce an answer or uses consistent units.

The workspace $W=(P,S)$ starts with no admitted findings in $S$. The
Orchestrator chooses assignments using the frontier $F(W)$: unresolved
plan nodes whose dependencies are given or admitted. Research outside
this frontier is logged, not blocked: a Researcher may investigate
unresolved prerequisites while answering its assigned subquestion.
Computations require available inputs. Here the plan stays unchanged
through 3 dispatches requesting
4, 3, and 1 research nodes.

\subsection{Brazuca: Findings, Feedback, and Evidence Completion}
\label{app:brazuca-example}

\noindent\textbf{First dispatch: obtain the runtime and identify the people.}
The initial assignments request the runtime, editor, production designer,
and director; the film target remains unresolved. Each Researcher first
uses a typed-resolver stage: focused retrieval, one extraction call, and
one verification attempt. Missing retrieval-hint inputs skip focused
retrieval. If no candidate is found or verification fails, the same
Researcher continues with a ReAct agent, retaining the candidate, sources,
and feedback.

The Researchers assigned runtime and director propose 19 minutes and
Faidon Gkretsikos at the typed-resolver stage. Their \emph{Brazuca} programme
page gives the production year, duration, director, birth year, Public
Administration studies, and third-short-film description. Both proposals
receive positive satisfaction and uniqueness judgments. Admission stores
$S(v)=(x,\mathcal{E}_v,\sigma)$: value, evidence references, and verification status.
These \textsc{verified} findings become available to later work, and the
runtime comparison evaluates to true. Admission changes $S$, not $P$.

\noindent\textbf{Continue research when the first lookup is insufficient.}
For editor and designer, the typed-resolver stage returns
\texttt{unsupported} without candidates; both Researchers continue in
ReAct within the same dispatch. The designer Researcher asks the separate
Reader whether the programme establishes both jobs. Its extraction with
supporting quotations reports ``Sets: Eleni Kariori'' but no art-director
credit. Reading Kariori's credits page then finds separate \emph{Brazuca}
(2017) entries under Production Designer and Art Director: film-specific
credits, not occupations inferred from the heading. The Researcher submits
Eleni Kariori with both passages. The Verifier endorses satisfaction and
uniqueness, and the harness admits the finding.

Search and reading supply material but do not resolve a plan node.
Before proposing a value, the ReAct Researcher can attach passages to
relations. Attachment checks scope and identifiers, not support for the
claim. The deterministic checks require attachments for checkable factual
relations whose partners are given or resolved; the Verifier assesses
whether the finding satisfies the assigned claim.
The initial retrieval stage sends its cited proposal directly to type
checking and LLM review, skipping the relation-attachment checks used in
ReAct, as shown in Figure~\ref{fig:paninidr-overview}.

\noindent\textbf{Use verification feedback to resolve an ambiguous identity.}
The editor Researcher submits Georgopoulos, citing the programme's
``Editing: Yorgos Georgopoulos'' and the editor list for \emph{The Other Me}.
The Verifier rejects the surname as non-unique, pointing to another listed
editor. The Researcher retrieves separate \emph{Brazuca} credits naming
Giorgos Georgopoulos and resubmits that full name with both films' passages,
now using the same spelling. Both judgments are positive: feedback leads
to a more specific proposal and sources establishing the shared-editor link.

The Verifier receives the plan node's ID and type, candidate, resolved
context, Researcher note, and passages. Known values fill the Planner's
verification claim; absent or unfilled claims fall back to relations
involving this node. ReAct supplies deduplicated proposal and
relation-bound citations, so evidence can be present even with no proposal
citations; the typed-resolver stage supplies its cited chunks. Endorsement
requires satisfaction and uniqueness; rejection explains missing support
or ambiguity. Missing decisions, unreadable replies, and failed calls do
not endorse the proposal. The Verifier does not search for rivals, treats snapshot dates
as given, and can defer clauses assigned elsewhere or awaiting unresolved
values. Its judgment covers the currently checkable claim.

\noindent\textbf{Second dispatch: use the admitted identities.}
With the people identified, Researchers seek the other film, the director's
film number, and his birth year. Their typed-resolver stages propose
\emph{The Other Me}, 3, and 1988. The other-film passage names Giorgos
Georgopoulos as editor and describes a criminology professor investigating
5 murders; the programme biography supplies the director values. All 3
receive positive satisfaction and uniqueness judgments. The remaining
comparisons evaluate to true: $3=3$, $1988<1990$, and $1988>1980$. All
7 intermediate facts and 4 \textsc{derived} comparisons are now stored,
but the target remains unresolved. Computations use admitted values without
another LLM check; the plan identifies inputs without a versioned history.

Bindings carry findings; retained research context helps later assignments
continue. ReAct keeps recent tool results in context and older ones as
retrievable pointers. Across assignments, a dossier for each researched subquestion retains
candidates, sources, feedback, and searches. Dossiers, conversations, and
archived results do not admit values: later research can use this context
without treating every note as an admitted finding.

\noindent\textbf{Third dispatch: complete the target's evidence.}
The target Researcher's typed-resolver stage returns \texttt{unsupported},
so it continues in ReAct. It binds the programme to the target's production-year,
runtime, director, editor, and production-design connections, then submits
\emph{Brazuca}. The deterministic checks refuse the update because the art-director
connection has not been bound. The Researcher reads Kariori's credits again
and attaches that passage to the missing connection. Its second submission
reaches the Verifier but is rejected: the supplied programme and designer
passages do not establish the connection to \emph{The Other Me}. After
another read of that film's editor credits, the Researcher submits
\emph{Brazuca} with all 3 passages. The Verifier now endorses satisfaction
and uniqueness, and the harness admits the target.

The workspace finishes with 8 factual bindings endorsed by the Verifier
and 4 \textsc{derived} Boolean outputs. The Orchestrator submits
\emph{Brazuca}. A separate LLM synthesis check assesses whether the final
answer follows from retrieved facts; its repair mode can replace the final
answer text with a high-confidence local correction. Here it accepts
\emph{Brazuca}, without warned admission or a Finalizer call. Ordinary
submission requires a resolved target, not a separate check that all
dependencies are resolved.

Separately, an LLM \emph{post-loop grounding check} assesses admitted
relations against cited passages or bounded article windows without changing bindings.
The plan's 2016 year matches
\emph{The Other Me (2016)}, but conflicts with the cited 19 January 2017
release date, which this check marks unsupported. The plot
clue guides retrieval but has no dedicated plan relation and is absent
from the final target Verifier's input.

\noindent\textbf{Verification guides further research.}
Across 3 target proposals in 1 dispatch, the title and plan stay fixed
while feedback prompts further reading: first to attach the art-director
evidence, then to establish the shared-editor link. Earlier findings guide
later assignments, and verification helps complete the evidence for the
target's checked conditions.

\subsection{When Research Takes a Different Path}
\label{app:other-execution-paths}

Brazuca returns one film, passes verification, and keeps its original plan.
Other investigations may seek several answers, exhaust verification
allowances, or reveal a missing research step.

\noindent\textbf{Answering from a collection of findings.}
MoNaCo \texttt{monaco\_1157} stores 3 government positions as one admitted
list, then computes the answer 3 with \texttt{count}. Like Brazuca's
comparisons, it stores a finding and then computes from it. Evidence and
verification cover the whole list, not each member; later admission replaces
the list rather than taking a union. Lists are not automatically complete:
the Verifier records completeness and missing items, but these fields do
not independently gate endorsement.

\noindent\textbf{Continuing after unsuccessful checks.}
Failed checks return feedback so the Researcher can gather more evidence,
as in Brazuca. The deterministic checks and the Verifier share a ReAct review cap of 5 failed-check
re-attempts, separate from the typed-resolver stage's single attempt. At
exhaustion, the harness can admit a \textsc{warned} value marked
\texttt{refuted\_best\_effort}. The plan node must remain known and
assigned, the value valid and correctly typed, and citation identifiers
valid. This admits without endorsement: incomplete evidence attachments skip the
Verifier, or admission follows a negative verdict.

The Orchestrator can retry unresolved research assignments with new hints and
retained context. To revisit an admitted finding, including a warned one,
it must request a supported plan revision that reopens the plan node,
then dispatch it again. Warnings do not automatically schedule retries.
Until reopened, verified and warned findings remain usable in research
and computation. Their verification flags remain in bindings, not ordinary
state or dispatch summaries, and \textsc{derived} outputs do not inherit
them. Appendix~\ref{app:ablation-verification} examines feedback and outcomes.

An assignment ends when its plan node resolves, its allowance runs out,
or the Researcher suggests revision, not through a separate self-reported
finish. Repeat assignments can have smaller allowances. The Orchestrator
can continue with assignments or revision until its overall allowance runs
out or repeated rounds without grounded progress trigger a stop.

\noindent\textbf{Adding a missing research step.}
BCP q1266 initially plans only the person being sought. Before dispatch,
the Orchestrator adds a subquestion for the series associated with that
person's 2016 breakthrough performance. This resolves on turn 2, the target
on turn 3, and turn 4 submits the correct answer, \emph{Maryam Tanveer Ali}.
Revision turns a missing intermediate fact into a research assignment
(Appendix~\ref{app:plan-revision-analysis}).

Only the Orchestrator requests updates; a Researcher's suggestion ends its
assignment without changing the plan. The Planner uses separate prompts
for initial plan creation and revision. The revision call receives the
question, current plan, bindings, reason, and evidence, and proposes edits
to the existing plan. The prompt asks for at most 2 requests;
the runtime permits 5, counting rejections, with up to 3 edit attempts per
request. At the 5-request limit, best-effort submission can bypass target
resolution. This occurred on 1 MoNaCo question and is separate from Finalizer
recovery.

Revision also determines which findings need reopening. A first pass
initializes new plan nodes, removes deleted ones, and resets surviving outputs
of added, removed, or changed computations. It follows \texttt{derived},
\texttt{argmax}/\texttt{argmin}, and comparison dependencies, not
\texttt{rank\_n}, \texttt{filter}, or \texttt{count}. A second pass reopens
preserved resolved nodes whose definitions changed and their
resolved descendants. First-pass resets are not starting points for this second pass,
so some affected findings remain resolved. Reopening clears values,
evidence, and verification records; a failed final structural check restores
the previous plan and state. Ordinary value replacement does not reset
descendants, and computations update only unresolved outputs. In BCP q591,
later edits leave an incorrect target resolved, preserving the wrong answer
(Appendix~\ref{app:plan-revision-analysis}).

\subsection{Fort Santiago: Answering from the Retained Research}
\label{app:example-record}

When research ends before resolving the target, the harness can make one
final LLM call to answer the original question from the retained record
without a new search. We refer to this fallback as the \emph{Finalizer}.
BCP q500 asks for a sixteenth-century structure's English name, using its
capital-city location, destruction, rebuilding, and later damage as clues.
Retained articles include \emph{Fort Santiago -- Wikipedia}, \emph{Welcome
to Fort Santiago!}, and \emph{About us}. The last page's retrieved text begins
``Welcome to Fort Santiago!'' and describes Legazpi re-establishing
Rajah Soliman's Manila in 1571: research has reached sources naming the
structure and describing its historical setting.

The difficulty is admitting the target. Across assignments, the
Researcher's typed-resolver stage proposes \texttt{1580} on 5 attempts;
the Verifier rejects a year in place of a name. ReAct also proposes
\emph{Fort Santiago} with citations, but 5 recorded updates stop at the deterministic checks: links to construction year, location, destruction, attacker,
and rebuilding lack attached passages. Citing the name does not complete
these bindings or establish the timeline, including the recorded
construction year 1580 and derived destruction year 1583.

After 5 consecutive rounds without grounded progress, the loop stops with
no answer. This triggers recovery: if read articles exist, the Finalizer
receives the question, resolved values, leading unresolved candidates, and
the 4 latest distinct articles, newest first. Limits are 45,000 characters
per article, 130,000 total, and 4,000 generated tokens, with no search or
read tools. If the answer remains empty, the \emph{stored-candidate fallback}
returns an eligible target candidate with evidence without another model call. Neither step
changes workspace bindings or verification judgments.

Here the Finalizer returns \emph{Fort Santiago}; all 3 grading judgments
accept it, and it passes Obs. The correct answer is recoverable from the
retained sources even though the target's relation-specific bindings remain
incomplete. Recovery makes that research useful without reversing the
earlier checks or marking the target verified
(Appendix~\ref{app:recovery-conditions}).

%% file: appendices/evidence_scoring.tex
Evidence bindings should leave a research record that supports checking
and reusing an answer. We first define how Obs. measures recoverable
answer credit, then test the matching procedure, contextual claim support,
answering from the collected evidence, and recorded calculations.

\noindent\textbf{What counts as evidence.}
The scorer uses saved search, read, and find outputs, source-validated
Reader quotations, and recorded inputs to the typed-resolver stage and
post-loop grounding check (Appendix~\ref{app:brazuca-example}). MoNaCo also includes recorded Reader windows.
IterResearch contributes page-summarizer inputs with HTML removed;
Open Deep Search contributes model-written search summaries. The agent's
own investigative prose and gold-document annotations are excluded.
The headline score also allows source text reconstructed from recorded
identifiers, including \paninidr{}'s references and Re-TRAC's self-citations.
Thus Obs. measures recoverability from the assembled record, not confirmed
receipt of every passage; Appendix~\ref{app:scorer-sensitivity} tests the
effect of removing identifier-only evidence.

\noindent\textbf{Combining correctness and evidence.}
On BCP and FRAMES, we test the judge-extracted answer, splitting lists,
dictionaries, and numeric/date compounds into checkable parts. Matching
uses fixed text, number, date, and name variants with token boundaries;
an all-but-one-token match requires at least 4 content tokens. Bare numbers
and short words require no additional proximity to question words, and
Boolean parts do not block support on these benchmarks. For official
score $o_i$, nonnegative credit $c_i=\max(0,o_i)$, answer parts
$\mathcal{P}_i$, and evidence $\mathcal{E}_i$, Equation~\ref{eq:observed-supported}
uses $s_i=\prod_{p\in\mathcal{P}_i}M(p,\mathcal{E}_i)$, where $M$ is the
fixed matching rule. A parsed answer with no items receives zero support.

\noindent\textbf{MoNaCo item-level credit.}
MoNaCo preserves partial answer credit. We select the judge pass with
highest official F1, breaking ties by Obs. and then pass order; its credited
items, prediction count, and support labels are used together. Let
$k_i^{\mathrm{obs}}$ count credited items that pass the support rule,
$g_i$ count gold items, and $n_i$ be the prediction count. We use the positive
judge count or, otherwise, the parsed-answer count, never below the credited
count. For multi-item questions,
\[
P_i^{\mathrm{obs}}=\frac{k_i^{\mathrm{obs}}}{\max(1,n_i,k_i^{\mathrm{obs}})},
\qquad R_i^{\mathrm{obs}}=\frac{\min(k_i^{\mathrm{obs}},g_i)}{g_i}.
\]
Let $H(P,R)=2PR/(P+R)$, or 0 when $P+R=0$. For a single-item question,
$s_i$ tests the judge-extracted answer. Then
\begin{equation}
\begin{gathered}
\widetilde{o}_i=
\begin{cases}
c_i s_i, & g_i=1,\\
\min\{c_i,H(P_i^{\mathrm{obs}},R_i^{\mathrm{obs}})\}, & g_i>1,\ n_i>0,\\
0, & \text{otherwise},
\end{cases}
\\[3pt]
\operatorname{Obs}_{\mathrm{MoNaCo}}=\frac{1}{N}\sum_{i=1}^{N}\widetilde{o}_i.
\end{gathered}
\label{eq:observed-supported-monaco}
\end{equation}
This caps recoverable credit at nonnegative official credit. MoNaCo's numeric official
scores can be negative; only Obs. floors them at zero. Empty and unjudged
answers receive zero support, with all 1,315 questions retained.

\noindent\textbf{Matching MoNaCo claims.}
We preserve the credited claim's numbers before normalizing its text.
Deterministic matching may use corresponding gold-item text to resolve
alternative wording, but cannot substitute a different number or require
unrelated gold attributes. A bare value is paired with its entity when
recoverable. Multi-part claims must match in one document, within 600
characters of a common anchor or with a text part identifying the document
through its title or first 400 normalized characters. Boolean-only claims
do not pass this base test.

%% file: appendices/scorer_sensitivity.tex
\subsection{Checking the Scorer and Evidence Record}
\label{app:scorer-sensitivity}

We test whether matching errors or reconstructed source text account for
the observed-support advantage. In a balanced development sample, separate
human and model reviews agreed on all 96 matching decisions: 8 passing
and 8 failing items from each combination of 3 systems (\paninidr{},
SLEUTH, Open Deep Search) and 2 benchmarks (BCP, MoNaCo). Review used the
matching source window, or the nearest window for a failed match.
MoNaCo consistency checks also reconcile credited and supported items
with the counts from the selected judge pass.

The fixed parser can discard credited answers: FRAMES has 6 such records
across systems---4 non-Latin answers, 1 Morse-code answer, and 1 null answer.
The normalization issue affects 2 \paninidr{} questions (0.24 points);
reported scores retain the fixed rules.

\noindent\textbf{Using only recorded evidence.}
Removing document bodies reconstructed solely from identifiers leaves all
7 assessable systems' BCP Obs. scores unchanged. The retained pool contains
saved tool outputs, source-validated Reader spans, and validated passages
shown to the typed-resolver stage and post-loop grounding check. \paninidr{}
remains at 70.7\% versus SLEUTH's 54.0\%, a paired gain of 16.7 points
(95\% interval $[+8.7,+24.7]$). The advantage therefore persists without
identifier-only evidence. Re-TRAC cannot be assessed because it saves
citations but no tool outputs.

\phantomsection
\noindent\textbf{Numerical-matching sensitivity.}
\label{app:derived-support}
As a secondary check, Obs.$^{+}$ allows unmatched numerical/Boolean parts
while still requiring text matches. On MoNaCo, it additionally credits
items mapped to numerical/Boolean gold values and recomputes F1, keeping
the selected pass and prediction count fixed. Denominators and official
judgments are unchanged. \paninidr{} remains highest on all 3 benchmarks
among available results; HyMem's MoNaCo Obs.$^{+}$ is unavailable.
Appendix~\ref{app:derivation-audit} complements this matching analysis by
checking recorded inputs and replaying calculations.

%% file: appendices/grounded_only_audit.tex
\subsection{Model-Based Grounded-Only Audit}
\label{app:grounded-only-audit}

To check whether sources support the requested claims, not merely contain
the answer text, we rebuild answers using only claims accepted by a common
model auditor. We compare \paninidr{} at cap 5, SLEUTH, and HyMem on all
150 BCP, 824 FRAMES, and 1,315 MoNaCo questions, using each system's own
saved evidence.

\texttt{gpt-5.6-luna} first splits answers into claims without judging
support. \texttt{gpt-5.6-terra} then checks them against the question and
evidence, using the same prompt and 80,000-character windows with 8,000
characters of overlap. The auditor sees neither gold answers nor official
verdicts. It accepts paraphrases but requires the requested attribute to
be supported in at least one window. We reconstruct answers from accepted
claims and re-grade them with \texttt{gpt-5.6-luna} under the shared official
protocol (Appendix~\ref{app:answer-grading}); empty reconstructions score
zero. Reconstructed answers containing all split claims provide a separate
formatting control in the analysis artifacts. Grounded-only reports the
official benchmark score of the reconstructed answer.

\input{tables/grounded_only_audit.tex}

HyMem can answer without retrieval, contributing to official scores but
not adding retrieved evidence for this audit (154 FRAMES, 139 MoNaCo, and
4 BCP questions). All questions nevertheless remain in the comparison.
\paninidr{} leads both baselines on all 3 benchmarks
(Table~\ref{tab:grounded-only-audit}): its advantage therefore extends from
recoverable answer text to answers rebuilt from claims judged supported
in context.

%% file: tables/grounded_only_audit.tex
\begin{table}[!htbp]
\centering
\small
\setlength{\tabcolsep}{7pt}
\begin{tabular}{lrrr}
\toprule
Method & BCP-150 & FRAMES-824 & MoNaCo-1315 \\
 & Accuracy (\%) & Accuracy (\%) & Mean-F1 \\
\midrule
HyMem & 48.0 & 43.6 & 0.352 \\
SLEUTH & 55.3 & 67.1 & 0.417 \\
\paninidr{} & \textbf{72.0} & \textbf{73.2} & \textbf{0.443} \\
\bottomrule
\end{tabular}
\caption{Answer quality after a common model auditor filters unsupported
claims. Rebuilt answers are re-graded under each benchmark's official
protocol. Every question is included, with empty answers scoring zero.
Bold marks the best score.}
\label{tab:grounded-only-audit}
\end{table}

%% file: appendices/fixed_reader_framing.tex
\subsection{Answering from Each System's Evidence}
\label{app:fixed-reader-framing}

To test whether retained research remains useful beyond the harness's own
answer generation, we give evidence from 4 systems to a common
\texttt{gpt-oss-120b} evaluation reader, with no new research or retrieval.
It generates new answers; the internal Reader extracts source evidence,
while Grounded-only filters original answers.

For each evidence set, the \emph{answer-present} prompt tells the reader
that the documents contain the answer and to use only those documents.
The \emph{uncertain-evidence} prompt instead warns that evidence may be
incomplete or irrelevant and allows abstention. Inputs and reader settings
are unchanged between conditions (temperature 0, seed 42). Each cell is one
run over all 150 BCP, 824 FRAMES, or 1,315 MoNaCo questions, scored with
\texttt{gpt-5.6-luna} under the shared 3-judgment protocol; empty answers
score zero. Each system supplies its collected evidence without matching
evidence volume or reading cost.

\input{tables/fixed_reader_framing.tex}

\paninidr{}'s evidence yields the highest answer quality on all 3 benchmarks
under both prompts (Table~\ref{tab:fixed-reader-framing}). With abstention
allowed, the lead over the strongest alternative is 62.7\% versus 55.3\%
on BCP, 69.4\% versus 62.5\% on FRAMES, and 0.547 versus 0.515 on MoNaCo.
The MoNaCo ordering against HyMem reverses: HyMem leads original-answer F1
(0.612 versus 0.549), but its evidence yields 0.419 versus our 0.547 here.
The retained material therefore remains useful beyond our own readout,
even when the reader is not promised that an answer is present.

\FloatBarrier

%% file: tables/fixed_reader_framing.tex
\begin{table}[!htbp]
\centering
\small
\setlength{\tabcolsep}{4pt}
\begin{tabular}{lrrrrrr}
\toprule
& \multicolumn{2}{c}{BCP-150 (\%)} & \multicolumn{2}{c}{FRAMES-824 (\%)} &
\multicolumn{2}{c}{MoNaCo-1315 (F1)} \\
\cmidrule(lr){2-3}\cmidrule(lr){4-5}\cmidrule(lr){6-7}
Evidence source & Present & Uncertain & Present & Uncertain & Present & Uncertain \\
\midrule
\paninidr{} & \textbf{72.0} & \textbf{62.7} & \textbf{76.7} & \textbf{69.4} & \textbf{0.601} & \textbf{0.547} \\
SLEUTH & 59.3 & 53.3 & 70.0 & 59.3 & 0.506 & 0.476 \\
HyMem & 54.7 & 50.0 & 52.2 & 43.9 & 0.470 & 0.419 \\
Omni-Decision & 58.7 & 55.3 & 71.2 & 62.5 & 0.538 & 0.515 \\
\bottomrule
\end{tabular}
\caption{Answer quality of a common evaluation reader using each system's saved evidence.
Present assumes the answer is available; Uncertain allows abstention.
All questions are included; empty answers score zero. Bold marks the best score.}
\label{tab:fixed-reader-framing}
\end{table}

%% file: appendices/obs_d.tex
\subsection{Checking Computed Answers}
\label{app:derivation-audit}

A computed answer need not appear in a source passage. We therefore test
whether the saved inputs and operations reproduce the answer, using Obs-D
to supplement text matching. This is a \paninidr{}-only audit: baseline
traces do not expose equivalent calculation chains.

For each target with positive official credit, we follow its recorded
calculation to given literals and retrieved inputs. Each retrieved input
must be nonempty and recoverable from received text in a document it cites;
empty collections do not establish support. Recorded quotations and inputs
to the typed-resolver stage and post-loop grounding check are source-validated;
citation identifiers alone cannot supply text. Matching pools recorded
text by cited document, rather than requiring the binding's exact chunk.
We then replay the recorded operations and compare the result with the
graded answer under fixed text, date, Boolean, and numeric matching rules,
without introducing an unrecorded conversion. A warned input excludes the
calculation; missing Verifier records are recorded as unknown, not accepted.

The Camp David example in Figure~\ref{fig:paninidr-overview} illustrates
the difference. The answer, 51, is absent from the 3 opened pages, so Obs.
gives no credit. Obs-D checks the recorded birth and signing dates against
received evidence and replays the completed-years operation to recover 51;
both date inputs have accepted Verifier records.

For a passing calculation, let $a_i$ be the nonnegative official credit
not already retained by Obs. Then
\begin{equation}
\operatorname{Obs\text{-}D}=\operatorname{Obs}
+\frac{1}{N}\sum_{i=1}^{N}a_i.
\label{eq:obs-d}
\end{equation}
FRAMES additions are binary; MoNaCo retains partial credit for the graded
derived item. Each passing target in these exports is its answer's sole
graded item. Fully covered answers receive no addition, and all questions
remain in the denominator. This automated audit checks whether recorded
calculations reproduce the graded answer from source-matched inputs.

\input{tables/derivation_audit.tex}

Replay adds credit on 51 FRAMES and 55 MoNaCo questions, raising scores
from 66.1\% to 72.3\% and from 0.458 to 0.498, respectively
(Table~\ref{tab:derivation-audit}). Every passing calculation has accepted
Verifier records for all retrieved inputs, so restricting additions to
that group leaves the gains unchanged. These answers extend the inspectable
record from facts found in text to results reproduced from those facts.

%% file: tables/derivation_audit.tex
\begin{table}[ht]
\centering
\small
\setlength{\tabcolsep}{7pt}
\begin{tabular}{lrr}
\toprule
 & FRAMES-824 & MoNaCo-1315 \\
\midrule
Arithmetic/Boolean target records & 192 & 335 \\
Passing calculations & 118 & 184 \\
Answers receiving additional credit & 51 & 55 \\
\midrule
Obs. & 66.1 & 0.458 \\
\textbf{Obs-D} & \textbf{72.3} & \textbf{0.498} \\
\bottomrule
\end{tabular}
\caption{\paninidr{}-only calculation audit. FRAMES scores are percentages;
MoNaCo scores are mean-F1. All questions remain in the denominator.
Only official credit not already retained by Obs. is added.}
\label{tab:derivation-audit}
\end{table}

%% file: appendices/additional_results.tex
\subsection{Research Evidence and Answer Completion}
\label{app:diagnostics}

\subsubsection{Finding Sources and Connecting Findings}
\label{app:evidence-coverage}
\label{app:matched-trajectories}

Persistent findings should help research reach relevant sources and retain
them for later use. We compare \paninidr{} and SLEUTH on matched questions,
measuring the fraction of annotated pages found and the coverage and
precision of pages recorded as support (Table~\ref{tab:evidence-coverage}).
\paninidr{} leads all three measures on every benchmark: broader discovery
is accompanied by a higher annotated share of retained support.
On BCP, its record contains every annotated page on 121 questions versus
46 for SLEUTH. Page presence includes snippets and recorded typed-resolver
and post-loop grounding inputs, not just full-page reads.

\input{tables/d_evidence_coverage.tex}

A matched trajectory shows how a finding can direct the next search.
BCP q276 asks which expedition an entrepreneur joined. \paninidr{}
identifies him as Jim Bridger, then uses his name to find a biography
connecting him to the Ashley--Henry expedition. It answers correctly
with observed support. SLEUTH never establishes the identity and returns
no answer. Here, persistent research state connects successive searches:
an intermediate answer supplies the next search subject.

\subsubsection{Revising the Plan and Recovering an Answer}
\label{app:plan-revision-analysis}
\label{app:recovery-analysis}
\label{app:recovery-conditions}
\label{app:admission-status}

Research can reveal a missing step in the initial plan. In the default
BCP and FRAMES executions, accepted revisions add 59 and 162 subquestions;
30 and 82 of these hold a value at the end. These counts describe new
research tasks and their completion, not a separate revision-on/off
accuracy effect. Appendix~\ref{app:other-execution-paths} follows an added
intermediate through to an answer and explains the rules for reopening
earlier findings.

When the target remains unfinished, answer recovery uses Finalizer synthesis
or stored-candidate fallback (Table~\ref{tab:finalizer}). On BCP, these paths supply 40 of the
110 correct answers, with 38 passing Obs. Recovery therefore preserves
useful research that would otherwise end without an answer; the Fort
Santiago trace in Appendix~\ref{app:example-record} shows this without
changing the workspace or reversing earlier verification judgments.
Final answer quality and workspace completion remain distinct. At the
default cap, warned findings comprise 2/666 (0.3\%), 31/2,638 (1.2\%),
and 235/3,619 (6.5\%) resolved nodes on BCP, FRAMES, and MoNaCo.

\input{tables/d_recovery_quality.tex}

\subsubsection{Answering with Annotated Sources}
\label{app:gold-reference}
\label{app:found-but-failed}

To test whether the same backbone can answer difficult questions once
relevant sources are supplied, we give \texttt{gpt-oss-120b} the annotated
documents without a search, planning, or verification loop.
The reference reaches 93.3\% BCP accuracy, including 45/50 hard questions
versus \paninidr{}'s 26/50 (Table~\ref{tab:gold-reference-results}).
This locates substantial room for improvement in finding and using
evidence, rather than answering ability alone.

\input{tables/d_gold_reference.tex}

Source packaging follows document volume. BCP includes up to 140,000
characters directly on 137 questions; the remaining 13 use an index and
up to nine 12,000-character windows. All 462 document entries resolve.
FRAMES includes up to 10 articles, each capped at 16,000 characters;
743 questions have clipped articles, retaining 26.7\% of selected text.
MoNaCo reads the first 60,000 characters of each resolvable page into
1,200-character notes, compressing notes above 40,000 characters before
answering. It resolves 40,606/56,734 entries and reads 71.5\% of their
text. Each cell is one execution, with all questions retained and the
same three-judgment aggregation as the headline scores. These different
input procedures make the reference a diagnostic, not a cost-matched
comparison or universal accuracy ceiling.

The reference also answers 14 of 17 BCP questions where \paninidr{}
found all annotated documents in returned tool records but answered
incorrectly. In q192, for example, research reaches the sources but
connects the target to the wrong artist's death. Finding the pages is
therefore only the first step: their facts must remain connected to the
right question through answer submission.

\subsection{Planning, Verification, and Recovery Controls}
\label{app:ablations}

\subsubsection{Structured Planning and Persistent State}
\label{app:ablation-workspace}

To test whether explicit research structure improves answers beyond prose
memory, we compare two reduced harnesses on all 150 BCP questions.
\textsc{Off} retains prose notes and Researcher reports;
\textsc{On} adds a plan, findings attached to its subquestions, and revision.
Both retain ReAct Researchers, direct source reading, and LLM verification
with five non-endorsement retries. Both omit the typed resolver, dedicated
Reader, executable calculations, relation-level deterministic checks, and auxiliary
full-system checks. \textsc{On} retains plan construction/repair and up to
five updates. The backbone, seed 42, temperature 0, and configured limits
are shared: 12 Orchestrator turns, 20 turns per Researcher, up to eight
Researchers per dispatch, and 32,000/16,384 prompt/completion tokens.
The intervention is this planning-and-state package, not a single component.

\input{tables/d_harness_control.tex}

\textsc{On} raises accuracy by 16.7 points: 37 answers become correct and
12 become incorrect (95\% paired interval $[+8.0,+25.3]$).
Hard-question successes double from 10 to 20. To check whether answer
recovery explains this advantage, we apply the same tool-free Finalizer
policy to each arm's empty-answer states. It receives the question, up to
four recently read articles, and workspace findings/candidates on
\textsc{On} or Researcher reports on \textsc{Off}. No \textsc{On} output
qualifies for stored-candidate fallback. The full system's recovery-off
cell withholds its 60 recovery outputs from the recorded run, without
rerunning research. Common recovery adds 23 correct answers to each
controlled arm, preserving the 16.7-point advantage
($[+9.3,+24.7]$). Thus, the gain persists when both versions can retain
notes, verify findings, and recover answers from their research.

With recovery, \textsc{On} matches the full system's 73.3\% BCP accuracy,
but the full system records fewer prompt tokens (1.43 M versus 1.78 M)
and higher Obs.\ (70.7 versus 68.7). Its Reader supplies focused source
extracts instead of placing full documents in the Researcher's context.
The typed resolver and computation together supply about half of the full
system's correct FRAMES answers and MoNaCo answers receiving credit.
The full package therefore offers benefits beyond matching BCP accuracy;
this comparison does not isolate the contributions of individual components.

We next check whether retained findings become inputs to later searches.
Among recorded entities absent from the question and followed by another
assignment, later searches reuse 192/248 (77.4\%) on \textsc{On} versus
176/548 (32.1\%) on \textsc{Off}. In q59, for example, \textsc{On} records
Maitreyi Raman and then searches for her award; \textsc{Off} encounters
the name but never searches with it. These entity matches show earlier
findings more often supplying the subject of subsequent research.
A stratified contextual check of 40 excerpts per arm labels 37 versus
28 genuine reuse; the original \textsc{On} annotations and two
\textsc{Off} annotations were manual, and an author reviewed the other
38 assistant-drafted \textsc{Off} labels. This check validates sampled
matches, not the causal contribution of reuse to accuracy.

\subsubsection{Verification Before Reuse}
\label{app:ablation-verification}

An incorrect intermediate finding can misdirect later research.
We therefore test the verification policy while retaining the workspace
and research tools. Off removes LLM review and the deterministic checks for valid citations and relation-specific source attachments. Assignment and type
checks and evidence logging remain. The on condition
uses the default cap of five. Both conditions use the same 150 BCP questions
and matched stratified samples of 200 FRAMES and 200 MoNaCo questions;
the MoNaCo on arm is the matching subset of the full cap-5 run.
Backbone, retrieval, seed, temperature, and scoring are shared.

\input{tables/d_verification.tex}

Full verification improves official and observed-support scores on all
three matched sets, adding 16.0 points to both BCP measures.
The gain includes research prompted by feedback, with median BCP prompt
use rising from 0.453 M to 1.042 M. The next analysis follows what
Researchers change before a previously rejected finding is accepted.

\subsubsection{Responding to Verification Feedback}
\label{app:ablation-cap}

The review cap controls failed-check re-attempts for an assigned plan node within
each dispatch, separately from the Researcher's turn allowance.
With the other full-system limits fixed, caps 1, 2, 5, and 10 on the same
150 BCP questions favor bounded review (Table~\ref{tab:cap-sweep}).
Cap 5 reaches 73.3\% accuracy with only two warned resolved findings;
cap 10 removes warnings but falls to 66.0\%. Cap 2 is close to cap 5,
whose paired accuracy advantage is 1.3 points with interval
$[-5.3,+8.0]$. More opportunities for verification therefore do not
consistently produce better answers.

\input{tables/cap_sweep.tex}

At cap 5, 77 of 101 Verifier-endorsed proposals following rejection
change the value or citation set relative to earlier rejected proposals:
40 introduce a new value and 37 pair an earlier value with a new citation
set. These are proposal events, excluding warned admissions and
evidence-attachment calls. The Brazuca editor sequence
(Appendix~\ref{app:brazuca-example}) shows the mechanism: feedback leads
to a full name and evidence connecting that person to both films.
Verification can thus guide further evidence gathering before reuse.

\subsection{Computation and Research Allowances}
\label{app:computation}

\subsubsection{Recorded Tokens and Cache Reuse}
\label{app:cost-profile}
\label{app:compute-cost}

\input{tables/research_cost.tex}

To explain the token expenditure in Table~\ref{tab:cost}, we break down
prompt use by role (Table~\ref{tab:role-cost}). The Researcher's ReAct
stage and Reader account for 85--89\% across benchmarks, versus 4--6\%
for the Orchestrator and 2--3\% for the LLM Verifier.

\input{tables/role_cost.tex}

On BCP, a question averages 109.45 ReAct model calls at 10.40k prompt
tokens each and 13.56 Reader requests at 9.92k prompt tokens per request.
A Reader request can include multiple model calls or a cache hit.
Deterministic checks require source attachments; the Verifier checks the
assigned claim and conditions. In the Brazuca trace, these checks
trigger further reading before admitting an unchanged title
(Appendix~\ref{app:brazuca-example}).

Table~\ref{tab:cost} uses recorded usage totals for IterResearch and
Re-TRAC, and counting proxies otherwise. Role sums aggregate calls
differently and do not partition trajectory totals.

To account for prefix-cache reuse, we measure engine counters in separate
runs of each harness over all 150 BCP questions: \texttt{vLLM}, two RTX PRO 6000
Blackwell GPUs, tensor parallelism 2, concurrency 8, seed 42, temperature
0, and one harness at a time after requests drain
(Table~\ref{tab:compute-cost}).

\input{tables/d_compute_cost.tex}

\paninidr{} submits 19.2 times SLEUTH's prompt tokens, but reuses 66.5\%
through the prefix cache. Uncached input plus generation gives a 7.3-fold
model-core computation ratio under
$\widehat F=2N_{\mathrm{active}}(T_{\mathrm{uncached\ input}}+T_{\mathrm{generated}})$,
with $N_{\mathrm{active}}=5.12\times10^9$.
It excludes attention and runtime overhead and is not GPU time or energy.
The following control compares repeated SLEUTH runs under this accounting.

\subsubsection{Repeated Sampling at Comparable Computation}
\label{app:matched-compute}

We give SLEUTH independent attempts to test whether repeated sampling can
match the complete harness. Each research run uses the headline configuration at
temperature 0.7 (our change from its greedy default), with seeds 42--49.
We select the most frequent non-empty normalized answer, breaking ties
by the lowest seed; both official and Obs. scores use the lowest-seed
run producing that answer and its evidence. If every answer is empty,
the question scores zero. Sampling runs use concurrency 16; engine counters
measure their actual cache reuse and cumulative computation.

\input{tables/d_matched_compute.tex}

At eight research runs per question, SLEUTH uses 1.12 times our model-core computation but
reaches 64.7\% accuracy versus 73.3\%; the paired advantage is 8.7 points
($[+1.3,+16.0]$) for both official and Obs. scores.
The rule fixed in advance continues past eight runs only if voted
accuracy is still increasing; it peaks at seven. Even selecting any
correct run would answer only 70.0\% of questions, although that is
an oracle choice, not the voting procedure. Under this tested sampling
policy, additional baseline attempts do not reproduce the harness's
answer quality.

\subsubsection{Extending Individual Investigations}
\label{app:ablation-budget}
\label{app:ablation-researcher-allowance}

Instead of restarting research, longer assignments can build on findings
already collected. We compare three allowance policies on all 150 BCP
questions (Table~\ref{tab:researcher-allowance-quality}). Increasing each
\paninidr{} Researcher's ReAct allowance from 20 to 50 and 80 turns,
with review cap 5 and the 12-turn Orchestrator limit fixed, raises accuracy
incrementally from 73.3\% to 74.7\% and 76.0\%. The 50-turn run gains
12 correct answers and loses 10; the 80-turn run gains 14 and loses 10.
The largest allowance uses 2.8 times the prompt tokens for 2.7 additional
points, favoring the lower-cost default for these executions.

\input{tables/d_allowances.tex}

The traces show additional turns completing support for existing candidates.
In q1204's 80-turn run, Stephan Weyte is proposed at turn 20; citation
repairs and further verification lead to admission at turn 26 without
changing the candidate. Across the separate 50- and 80-turn executions,
24 and 26 questions move from a correct recovery answer at the default
to a correct answer with a resolved workspace target. This shift is
larger than the net accuracy gain: longer research can complete evidence
that recovery already used to answer.

The shared-budget panel instead limits combined Orchestrator/Researcher
turns. Remaining turns are split across dispatched Researchers with a
minimum of one each, so a final dispatch can exceed the nominal cap.
The default has separate limits, not a shared question budget; its higher
accuracy accompanies higher expenditure. SLEUTH's own allowance sweep
shows no sustained gain: 57.3\%, 59.3\%, and 57.3\%, despite mean prompt
use rising from 74k to 111k. Input-length refusals affect 19, 18, and 26
of 150 questions, respectively, while total unanswered counts are 26, 22,
and 28. From 20 to 80 turns, budget exhaustion falls from 7 to 1 question,
so answer coverage decreases by only 2 questions despite 7 additional
input-length refusals.
All unanswered questions remain in the scores. These turn policies cover
different activity; only the preceding sampling control compares
model-core computation.

\subsection{Run Stability, Paired Comparisons, and Answer Structure}
\label{app:uncertainty}

\phantomsection
\noindent\textbf{Stability across research runs.}
\label{app:run-stability}
To check stability, we evaluate three research runs per harness under the
headline BCP settings, including the headline run
(Table~\ref{tab:bcp-run-stability}). \paninidr{} averages 72.7\% accuracy,
versus 58.0\% for both SLEUTH and Open Deep Search---a 14.7-point lead.
Its lowest-scoring run exceeds every baseline run in this comparison.

\input{tables/bcp_run_stability.tex}

\noindent\textbf{Paired question-level comparisons.}
To check whether reported gains depend on the sampled questions,
Table~\ref{tab:key-uncertainty} gives 95\% intervals from 10,000 paired
bootstrap resamples of unrounded question-level differences. All
questions keep their benchmark weight, including failures with zero
credit. These intervals describe question-sampling uncertainty within
the recorded executions, not variation across repeated research runs.
The listed Obs. advantages over baselines remain positive, while several
official-score differences include zero or favor another harness.

On FRAMES, the accuracy advantage over SLEUTH has a paired
95\% interval that includes zero ($[-1.0,+4.9]$;
Table~\ref{tab:key-uncertainty}). On MoNaCo, \paninidr{} trails
HyMem, Open Deep Search, IterResearch, and SLEUTH in overall F1; on collection questions, its highest
precision (0.690) does not compensate for lower recall (0.462 versus
HyMem's 0.566; Table~\ref{tab:monaco-decomposition}).

Because SLEUTH's input-length refusals could partly explain the BCP gap,
we exclude the same 19 questions from both systems. On the remaining
131, accuracy is 74.8\% versus 65.6\%, a gain of 9.2 points
($[+0.8,+17.6]$); Obs. is 71.8\% versus 61.8\%, a gain of 9.9 points
($[+1.5,+18.3]$). The advantage therefore persists on this common subset.
This removes recorded refusals, rather than predicting performance with
a different context window.

\input{appendices/monaco_decomposition.tex}

\input{tables/d_paired_comparisons.tex}

\subsection{Baseline Adaptations}
\label{app:compat}

Table~\ref{tab:compat} separates
paper-based reimplementations from released-code adaptations. SLEUTH,
Omni-Decision, and HyMem had no available code when experiments were
established; we follow their published specifications rather than
reproducing released implementations.

\input{tables/d_baseline_compatibility.tex}

Omni-Decision leaves the zero-novelty termination limit unspecified.
We use three, the least restrictive setting tested: limits
one, two, and three lose six, two, and zero answers on FRAMES-50.
The rule fires on 433/1,315 MoNaCo questions; its effect there is unmeasured.
Re-TRAC uses its repository entry point and unchanged configuration file
with the shared backbone, not its specialized fine-tuned checkpoint.
On BCP it produces 76 empty outputs and seven abstentions: its final node
returns the last trajectory message, which can be a tool call rather
than answer text. These 83 outcomes score zero in the reported comparison.

%% file: tables/d_evidence_coverage.tex
\begin{table}[!htbp]
\centering\small
\setlength{\tabcolsep}{6pt}
\begin{tabular}{lrrrr}
\toprule
Benchmark & $n$ & Found recall & Cited recall & Cited precision (\%) \\
\midrule
BCP & 150 & .860 / .455 & .485 / .248 & 43.3 / 37.0 \\
FRAMES & 823 & .945 / .820 & .774 / .586 & 91.9 / 84.4 \\
MoNaCo & 1,315 & .515 / .368 & .281 / .250 & 73.3 / 67.0 \\
\bottomrule
\end{tabular}
\caption{Source coverage: \paninidr{} / SLEUTH. Recall is the fraction
of annotated pages found or recorded as support; precision is the
annotated share of support pages. Values average per-question fractions,
with zero precision when no citation is recorded. Source measures omit
one FRAMES question without annotations; answer scores retain all 824.
SLEUTH source titles are normalized and matched to recorded pages.}
\label{tab:evidence-coverage}
\end{table}

%% file: tables/d_recovery_quality.tex
\begin{table}[!htbp]
\centering\small
\begin{tabular}{lrrr}
\toprule
Benchmark & Recovery answers & Official & Obs. \\
\midrule
BCP & 60 & 66.7 & 63.3 \\
FRAMES & 103 & 60.2 & 52.4 \\
MoNaCo & 250 & 0.430 & 0.358 \\
\bottomrule
\end{tabular}
\caption{Scores over all identified recovery answers, including incorrect
ones: percentages on BCP/FRAMES and mean F1 on MoNaCo. Finalizer outputs
number 50/89/241 and stored-candidate fallback outputs 10/14/9, respectively.
One additional MoNaCo output has an unclassified origin.}
\label{tab:finalizer}
\end{table}

%% file: tables/d_gold_reference.tex
\begin{table}[!htbp]
\centering\small
\begin{tabular}{lrrrrrr}
\toprule
 & \multicolumn{4}{c}{BCP} & FRAMES & MoNaCo \\
\cmidrule(lr){2-5}
 & Easy & Med. & Hard & All & Acc. & F1 \\
\midrule
\paninidr{} & 92 & 76 & 52 & 73.3 & 79.1 & 0.549 \\
Annotated-source reference & 96 & 94 & 90 & 93.3 & 79.5 & 0.550 \\
\bottomrule
\end{tabular}
\caption{Official scores with research or supplied sources: accuracy (\%)
on BCP/FRAMES and mean F1 on MoNaCo. Denominators are 150/824/1,315;
each BCP difficulty group contains 50 questions.}
\label{tab:gold-reference-results}
\end{table}

%% file: tables/d_harness_control.tex
\begin{table}[!htbp]
\centering\small
\setlength{\tabcolsep}{5pt}
\begin{tabular}{lrrrrr}
\toprule
 & \multicolumn{2}{c}{Recovery off} & \multicolumn{2}{c}{Recovery on} & Prompt (M) \\
\cmidrule(lr){2-3}\cmidrule(lr){4-5}
Arm & Official & Obs. & Official & Obs. & All roles, mean \\
\midrule
\textsc{Off}: prose memory & 41.3 & 39.3 & 56.7 & 53.3 & 1.38 \\
\textsc{On}: planning/state & 58.0 & 56.0 & 73.3 & 68.7 & 1.78 \\
Full system & 46.7 & 45.3 & 73.3 & 70.7 & 1.43 \\
\bottomrule
\end{tabular}
\caption{Planning/state and common-recovery controls on all 150 BCP
questions; scores are percentages. Prompt means cover the recorded
reference runs: they exclude post-hoc Finalizer calls on the controlled
arms and include live recovery in the full system. Full-system rows
restore multiple components and are not isolated planning/state controls.}
\label{tab:harness-cot-control}
\label{tab:recovery-control}
\end{table}

%% file: tables/d_verification.tex
\begin{table}[!htbp]
\centering\small
\setlength{\tabcolsep}{6pt}
\begin{tabular}{llrrr}
\toprule
Benchmark & Verification & Official & Obs. & Prompt (k) \\
\midrule
BCP (150) & On & 73.3 & 70.7 & 1,042 \\
 & Off & 57.3 & 54.7 & 453 \\
FRAMES (200) & On & 79.5 & 69.0 & 125 \\
 & Off & 69.5 & 57.0 & 35 \\
MoNaCo (200) & On & 0.583 & 0.479 & 288 \\
 & Off & 0.517 & 0.412 & 47 \\
\bottomrule
\end{tabular}
\caption{Verification ablation on matched questions. Off removes LLM review
and source-evidence checks but retains assignment and type checks. Scores are percentages on
BCP/FRAMES and mean F1 on MoNaCo. Prompt use is the per-question median
in thousands, using trajectory accounting.}
\label{tab:llm-verifier-ablation}
\end{table}

%% file: tables/cap_sweep.tex
\begin{table}[!htbp]
\centering
\small
\setlength{\tabcolsep}{5pt}
\begin{tabular}{rrrrrr}
\toprule
Cap & Acc. & Obs. & Hard / 50 & Warned / resolved & Prompt (M) \\
\midrule
1 & 67.3 & 65.3 & 19 & 128 / 773 & 1.249 (0.817) \\
2 & 72.0 & 68.7 & 23 & 27 / 636 & 1.147 (0.854) \\
\textbf{5} & \textbf{73.3} & \textbf{70.7} & \textbf{26} & 2 / 666 & 1.200 (1.042) \\
10 & 66.0 & 63.3 & 20 & 0 / 673 & 1.317 (1.115) \\
\bottomrule
\end{tabular}
\caption{Shared review-cap sensitivity on BCP CF-150. Accuracy and Obs.
are percentages; Hard counts correct answers out of 50. Warning counts
use resolved bindings in the final workspace snapshots, including computed
bindings. Prompt use is the per-question mean in millions, with the median
in parentheses, using the trajectory accounting in Table~\ref{tab:cost}.
These are benchmark averages for the supplied executions, not averages
across research repetitions.}
\label{tab:cap-sweep}
\end{table}

%% file: tables/research_cost.tex
\begin{table}[!htbp]
\centering
\small
\setlength{\tabcolsep}{8pt}
\begin{tabular}{lrr}
\toprule
\textbf{Method} & \textbf{Prompt (M)} & \textbf{Completion (k)} \\
\midrule
PyRAG                  & 0.002 (0.017) & 3.2 (6.5) \\
SLEUTH                 & 0.037 (0.074) & 2.3 (7.3) \\
Open Deep Search       & 0.10 (0.11)   & 10.0 (10.9) \\
Omni-Decision          & 0.13 (0.16)   & 9.0 (10.1) \\
IterResearch           & 0.17 (0.17) & 9.8 (10.6) \\
HyMem                  & 0.29 (0.31)   & 51.0 (65.9) \\
Re-TRAC                & 5.12 (5.34)   & 33.6 (42.7) \\
\midrule
\textbf{\paninidr{}}   & 1.04 (1.20)   & 36.1 (44.2) \\
\bottomrule
\end{tabular}

\caption{BCP CF-150 token use: per-question medians (means), with
prompt tokens in millions and completion in thousands. All rows use
\texttt{gpt-oss-120b}. Counting details appear in
Appendix~\ref{app:cost-profile}.}
\label{tab:cost}
\end{table}

%% file: tables/role_cost.tex
\begin{table}[!htbp]
\centering
\small
\setlength{\tabcolsep}{4pt}
\begin{tabular}{lrrr}
\toprule
 & BCP & FRAMES & MoNaCo \\
\midrule
Questions & 150 & 824 & 1,315 \\
Reported trajectory prompt total (M) & 180.04 & 270.38 & 775.12 \\
Sum of 10 recorded roles (M prompt tokens) & 213.94 & 340.79 & 1,040.13 \\
Role sum / trajectory total & 1.188 & 1.260 & 1.342 \\
\midrule
\multicolumn{4}{l}{Selected roles: mean prompt tokens per question, k (share of role sum)} \\
Orchestrator & 58.4 (4.1\%) & 23.1 (5.6\%) & 32.8 (4.1\%) \\
Researcher (ReAct) & 1,138.0 (79.8\%) & 303.7 (73.4\%) & 550.8 (69.6\%) \\
Reader & 134.6 (9.4\%) & 48.8 (11.8\%) & 156.2 (19.7\%) \\
LLM Verifier & 40.5 (2.8\%) & 7.5 (1.8\%) & 13.7 (1.7\%) \\
ReAct + Reader share (\%) & 89.2 & 85.2 & 89.4 \\
\bottomrule
\end{tabular}
\caption{\paninidr{} cap-5 role accounting from saved traces.
Role telemetry covers every question; shares use summed role prompt
tokens across all 10 roles, including the 6 not listed individually.
ReAct denotes the Researcher's second stage, excluding the typed resolver.
k and M denote thousands and millions of tokens. Combined shares use
unrounded counts. Table~\ref{tab:cost} instead reports trajectory totals.}
\label{tab:role-cost}
\end{table}

%% file: tables/d_compute_cost.tex
\begin{table}[!htbp]
\centering\small
\setlength{\tabcolsep}{5pt}
\begin{tabular}{lrrrrr}
\toprule
Harness & Input (k) & Cache (\%) & Uncached (k) & Generated (k) & PFLOP \\
\midrule
\paninidr{} & 1,422 & 66.5 & 477 & 78 & 5.68 \\
SLEUTH & 74 & 7.3 & 69 & 7 & 0.78 \\
Omni-Decision & 157 & 7.1 & 146 & 10 & 1.59 \\
HyMem & 309 & 45.0 & 170 & 66 & 2.41 \\
\bottomrule
\end{tabular}
\caption{Engine measurements over all 150 BCP questions. Tokens and
analytically estimated model-core PFLOPs are per-question means;
cache percentage is the share of submitted input reused.}
\label{tab:compute-cost}
\end{table}

%% file: tables/d_matched_compute.tex
\begin{table}[!htbp]
\centering\small
\setlength{\tabcolsep}{5pt}
\begin{tabular}{rrrrrr}
\toprule
Runs & PFLOP/q & vs. ours & Vote acc. & Any correct & Vote Obs. \\
\midrule
1 & 0.76 & 0.13 & 61.3 & 61.3 & 58.7 \\
2 & 1.66 & 0.29 & 62.0 & 66.0 & 59.3 \\
4 & 3.22 & 0.57 & 63.3 & 69.3 & 60.7 \\
6 & 4.79 & 0.84 & 64.7 & 69.3 & 61.3 \\
7 & 5.51 & 0.97 & 65.3 & 70.0 & 62.7 \\
8 & 6.34 & 1.12 & 64.7 & 70.0 & 62.0 \\
\midrule
\multicolumn{3}{l}{\paninidr{}: 5.68 PFLOP/q} & 73.3 & --- & 70.7 \\
\bottomrule
\end{tabular}
\caption{Repeated SLEUTH research runs on all 150 BCP questions at cumulative
model-core computation. Vote selects the most frequent non-empty answer;
``Any correct'' credits a question if at least one run answers correctly.
Scores are percentages. Individual runs average 55.5\% accuracy
versus the greedy reference's 57.3\%.}
\label{tab:matched-compute}
\end{table}

%% file: tables/d_allowances.tex
\begin{table}[!htbp]
\centering\small
\setlength{\tabcolsep}{6pt}
\begin{tabular}{lrrrr}
\toprule
Allowance & Accuracy & Obs. & Mean prompt (M) & Empty \\
\midrule
\multicolumn{5}{l}{\paninidr{}: turns per Researcher, review cap 5} \\
20 (default) & 73.3 & 70.7 & 1.20 & 4 \\
50 & 74.7 & 71.3 & 2.54 & 2 \\
80 & 76.0 & 74.0 & 3.33 & 6 \\
\midrule
\multicolumn{5}{l}{\paninidr{}: shared question budget, review cap 5} \\
20 & 31.3 & 28.7 & 0.115 & 71 \\
50 & 49.3 & 46.7 & 0.279 & 38 \\
80 & 57.3 & 54.0 & 0.440 & 34 \\
\midrule
\multicolumn{5}{l}{SLEUTH: its own turn allowance} \\
20 & 57.3 & 54.0 & 0.074 & 26 \\
50 & 59.3 & 56.0 & 0.106 & 22 \\
80 & 57.3 & 54.0 & 0.111 & 28 \\
\bottomrule
\end{tabular}
\caption{Separate allowance sweeps on all 150 BCP questions, including
unanswered cases. Scores are percentages; prompt means use trajectory
accounting. Turn limits govern different activity across panels and are
not matched computation. Each row is a separate execution, not a
continuation of the shorter run.}
\label{tab:researcher-allowance-quality}
\label{tab:shared-turn-budget-sensitivity}
\end{table}

%% file: tables/bcp_run_stability.tex
\begin{table}[!htbp]
\centering\small
\setlength{\tabcolsep}{5pt}
\begin{tabular}{lrrrrr}
\toprule
Method & Table~\ref{tab:main-results} & Run 2 & Run 3 & Mean $\pm$ SD & Mean (\%) \\
\midrule
\textbf{\paninidr{}} & \textbf{110} & \textbf{107} & \textbf{110} & $\mathbf{109.0\pm1.7}$ & \textbf{72.7} \\
SLEUTH & 86 & 91 & 84 & $87.0\pm3.6$ & 58.0 \\
Open Deep Search & 80 & 88 & 93 & $87.0\pm6.6$ & 58.0 \\
IterResearch & 82 & 83 & 80 & $81.7\pm1.5$ & 54.4 \\
HyMem & 73 & 73 & 74 & $73.3\pm0.6$ & 48.9 \\
Omni-Decision & 71 & 68 & 59 & $66.0\pm6.2$ & 44.0 \\
Re-TRAC & 59 & 52 & 52 & $54.3\pm4.0$ & 36.2 \\
PyRAG & 41 & 48 & 43 & $44.0\pm3.6$ & 29.3 \\
\bottomrule
\end{tabular}
\caption{BCP stability across three research runs per
method, each on the same 150 questions under the headline settings.
Run columns and mean $\pm$ sample SD count correct answers; the final
column converts the mean count to accuracy. The first run is the one
reported in Table~\ref{tab:main-results}.}
\label{tab:bcp-run-stability}
\end{table}

%% file: appendices/monaco_decomposition.tex
\noindent\textbf{MoNaCo precision, recall, and answer structure.}
To explain the MoNaCo accuracy trade-off, we separate finding one answer
from enumerating a collection. Table~\ref{tab:monaco-decomposition} groups
the 720 single-item and 595 collection questions by the official judge's
gold-item counts. Precision and recall use the same selected maximum-F1
pass as the headline score. Each class retains all its questions, with
missing predictions scoring zero.

\input{tables/monaco_decomposition.tex}

On collections, \paninidr{} has the highest precision (0.690), but its
0.462 recall leaves F1 at 0.495: finding fewer wrong items does not
compensate for missing valid ones. Against SLEUTH, it scores higher
on single-item F1 (0.593 versus 0.579) but lower on collection F1
(0.495 versus 0.528). \paninidr{} leads overall and on single-item
observed support; its collection Obs. is 0.474.

%% file: tables/monaco_decomposition.tex
\begin{table}[htbp]
\centering
\small
\setlength{\tabcolsep}{4pt}
\begin{tabular}{lrrrrrr}
\toprule
 & \multicolumn{2}{c}{Single item (720)} &
 \multicolumn{4}{c}{Collection (595)} \\
\cmidrule(lr){2-3}\cmidrule(lr){4-7}
Harness & F1 & Obs. & P & R & F1 & Obs. \\
\midrule
\paninidr{} & 0.593 & \textbf{0.446} & \textbf{0.690} & 0.462 & 0.495 & 0.474 \\
SLEUTH & 0.579 & 0.338 & 0.678 & 0.503 & 0.528 & 0.467 \\
Open Deep Search & 0.598 & 0.269 & 0.652 & 0.519 & 0.538 & 0.394 \\
IterResearch & 0.583 & 0.317 & 0.617 & 0.546 & 0.540 & \textbf{0.479} \\
Omni-Decision & 0.410 & 0.282 & 0.399 & 0.281 & 0.298 & 0.267 \\
Re-TRAC & 0.572 & 0.360 & 0.591 & 0.507 & 0.511 & 0.441 \\
PyRAG & 0.474 & 0.297 & 0.581 & 0.380 & 0.413 & 0.375 \\
HyMem & \textbf{0.651} & 0.360 & 0.654 & \textbf{0.566} & \textbf{0.566} & 0.423 \\
\bottomrule
\end{tabular}
\caption{MoNaCo answer structure for the main-table executions.
Columns average over all questions in each group using the same maximum-F1
pass. Mean F1 is not the harmonic mean of aggregate precision and recall.
Bold marks the largest value.}
\label{tab:monaco-decomposition}
\end{table}

%% file: tables/d_paired_comparisons.tex
\begin{table}[!htbp]
\centering
\footnotesize
\setlength{\tabcolsep}{3pt}
\begin{tabular}{llrrcrc}
\toprule
& & & \multicolumn{2}{c}{Official} & \multicolumn{2}{c}{Obs.} \\
\cmidrule(lr){4-5}\cmidrule(lr){6-7}
Dataset & Comparator & $n$ & $\Delta$ & 95\% CI & $\Delta$ & 95\% CI \\
\midrule
\multicolumn{7}{l}{\textit{Headline comparisons: }\paninidr{}\textit{ (cap 5) minus baseline}} \\
BCP & SLEUTH & 150 & $+16.0$ & $[+8.0, +24.0]$ & $+16.7$ & $[+8.7, +24.7]$ \\
FRAMES & SLEUTH & 824 & $+1.9$ & $[-1.0, +4.9]$ & $+10.9$ & $[+7.8, +14.1]$ \\
MoNaCo & SLEUTH & 1,315 & $-0.007$ & $[-0.029, +0.015]$ & $+0.062$ & $[+0.041, +0.083]$ \\
BCP & HyMem & 150 & $+24.7$ & $[+16.0, +34.0]$ & $+24.0$ & $[+14.7, +32.7]$ \\
FRAMES & HyMem & 824 & $+8.4$ & $[+5.0, +11.7]$ & $+29.5$ & $[+25.7, +33.1]$ \\
MoNaCo & HyMem & 1,315 & $-0.064$ & $[-0.086, -0.042]$ & $+0.070$ & $[+0.046, +0.095]$ \\
MoNaCo & Re-TRAC & 1,315 & $+0.004$ & $[-0.021, +0.029]$ & $+0.062$ & $[+0.038, +0.086]$ \\
MoNaCo & Open Deep Search & 1,315 & $-0.022$ & $[-0.044, -0.001]$ & $+0.133$ & $[+0.110, +0.156]$ \\
MoNaCo & IterResearch & 1,315 & $-0.015$ & $[-0.040, +0.009]$ & $+0.068$ & $[+0.043, +0.094]$ \\
MoNaCo & PyRAG & 1,315 & $+0.102$ & $[+0.079, +0.127]$ & $+0.126$ & $[+0.105, +0.148]$ \\
BCP & SLEUTH, non-refused & 131 & $+9.2$ & $[+0.8, +17.6]$ & $+9.9$ & $[+1.5, +18.3]$ \\
\midrule
\multicolumn{7}{l}{\textit{BCP cap sweep: cap 5 minus comparator}} \\
BCP & Cap 1 & 150 & $+6.0$ & $[-1.3, +13.3]$ & $+5.3$ & $[-2.0, +12.7]$ \\
BCP & Cap 2 & 150 & $+1.3$ & $[-5.3, +8.0]$ & $+2.0$ & $[-4.7, +9.3]$ \\
BCP & Cap 10 & 150 & $+7.3$ & $[+0.7, +14.0]$ & $+7.3$ & $[+0.7, +14.0]$ \\
\midrule
\multicolumn{7}{l}{\textit{BCP shared budget at cap 5: hierarchical reference minus shared budget}} \\
BCP & Shared 80 & 150 & $+16.0$ & $[+8.0, +24.0]$ & $+16.7$ & $[+8.7, +24.7]$ \\
BCP & Shared 50 & 150 & $+24.0$ & $[+15.3, +32.7]$ & $+24.0$ & $[+15.3, +32.7]$ \\
BCP & Shared 20 & 150 & $+42.0$ & $[+33.3, +50.7]$ & $+42.0$ & $[+33.3, +50.7]$ \\
\midrule
\multicolumn{7}{l}{\textit{Verification ablation: verification on (cap 5) minus verification off}} \\
BCP & Verification off & 150 & $+16.0$ & $[+8.0, +24.0]$ & $+16.0$ & $[+8.0, +24.0]$ \\
FRAMES & Verification off & 200 & $+10.0$ & $[+3.5, +16.5]$ & $+12.0$ & $[+5.5, +18.5]$ \\
MoNaCo & Verification off & 200 & $+0.066$ & $[+0.015, +0.118]$ & $+0.067$ & $[+0.019, +0.116]$ \\
\midrule
\multicolumn{7}{l}{\textit{Planning and state: }\textsc{On}\textit{ minus }\textsc{Off}} \\
BCP & Planning/state \textsc{Off} & 150 & $+16.7$ & $[+8.0, +25.3]$ & $+16.7$ & $[+8.0, +25.3]$ \\
\midrule
\multicolumn{7}{l}{\textit{Recovery control: +R marks the full system's Finalizer applied to a controlled arm}} \\
BCP & \textsc{On}+R $-$ \textsc{Off}+R & 150 & $+16.7$ & $[+9.3, +24.7]$ & $+15.3$ & $[+7.3, +23.3]$ \\
BCP & \textsc{On}+R $-$ \textsc{On} & 150 & $+15.3$ & $[+10.0, +21.3]$ & $+12.7$ & $[+7.3, +18.0]$ \\
BCP & \textsc{Off}+R $-$ \textsc{Off} & 150 & $+15.3$ & $[+10.0, +21.3]$ & $+14.0$ & $[+8.7, +20.0]$ \\
BCP & Full $-$ \textsc{On}+R & 150 & $+0.0$ & $[-6.7, +7.3]$ & $+2.0$ & $[-4.7, +9.3]$ \\
\midrule
\multicolumn{7}{l}{\textit{Matched computation: }\paninidr{}\textit{ minus 8-sample SLEUTH at 1.12 times our PFLOP}} \\
BCP & 8-sample SLEUTH & 150 & $+8.7$ & $[+1.3, +16.0]$ & $+8.7$ & $[+1.3, +16.0]$ \\
\bottomrule
\end{tabular}

\caption{Paired score differences on the matched question sets, with
bootstrap 95\% confidence intervals from 10,000 resamples paired by
question. Positive differences favor the first-named arm in each
subtraction. Values are percentage points on BCP and FRAMES and
mean-F1 units on MoNaCo; $n$ is the number of matched questions.
Differences use unrounded scores. Each comparison resamples the same
question set in both arms, so an interval describes the difference rather
than either score separately.}
\label{tab:key-uncertainty}
\label{tab:sleuth-nonrefused}
\end{table}

%% file: tables/d_baseline_compatibility.tex
\begin{table}[htbp]
\centering\small
\begin{tabularx}{\linewidth}{@{}lX@{}}
\toprule
System & Implementation or adaptation \\
\midrule
SLEUTH & Paper/prompt-based: published reasoning, memory, commitment, and top-3 retrieval rules; dense and BM25 routes. \\
Omni-Decision & Paper-based: 15 iterations, top-6 retrieval, zero-novelty termination after 3 iterations. \\
HyMem & Paper-based: 4 tools, published tool wording, and per-layer budgets. \\
Open Deep Search & Released code: search timeout raised from 10 to 300 s; local-link gateway. \\
PyRAG & Output limit raised from 1,024 to 4,096 tokens; JSON-schema output. \\
IterResearch & Endpoint/return-value compatibility; assistant prefill, XML repair, JSON lift, action seeding, answer extraction; authors' hard-coded temperature 0.6. \\
Re-TRAC & Repository prompting configuration; endpoint/return-value compatibility and a persistent driver event loop. \\
\bottomrule
\end{tabularx}
\caption{Baseline implementations and changes for the shared evaluation
environment. Model-facing changes are listed separately from interface
compatibility within each row.}
\label{tab:compat}
\end{table}

%% file: appendices/related_work.tex
\section{Extended Related Work}
\label{app:related-work}

The methods below address complementary parts of a research investigation:
how work is decomposed, what persists between calls, how findings are
checked, and how answers are evaluated. We first situate the representation,
then explain the research harnesses individually and connect their mechanisms
to verification and evidence assessment.

\subsection{Reasoning Structure, Plans, and Semantic Workspaces}
\label{app:related-plans}

\noindent\textbf{Reasoning chains and search.}
Chain-of-thought prompting elicits intermediate steps before an answer
\citep{wei2022chainofthought}. Self-consistency samples diverse reasoning
paths and selects an answer by aggregating across them
\citep{wang2023selfconsistency}. Tree of Thoughts explores alternative
reasoning paths, and Graph of Thoughts supports more general combinations of
intermediate results \citep{yao2023treeofthoughts,besta2024graphofthoughts}.
These methods organize how reasoning is generated, searched, or combined.
\paninidr{} addresses its continuation across research calls: admitted
intermediate values determine what subsequent assignments can use.

\noindent\textbf{Follow-up questions and retrieval.}
Self-Ask explicitly asks and answers follow-up questions before answering
the original query, and can use a search engine to supply the intermediate
answers \citep{press2023selfask}. IRCoT lets intermediate reasoning guide
retrieval and retrieved passages inform subsequent reasoning
\citep{trivedi2023ircot}. ReAct interleaves reasoning, actions, and observations
to update a course of action through tool interaction \citep{yao2023react}.
These methods connect decomposition to evidence gathering. In \paninidr{},
the research subquestion persists beyond an individual interaction: it
specifies the assignment, the claim to check, and where an admitted answer
enters the shared state.

\noindent\textbf{Executable intermediate reasoning.}
Faithful Chain-of-Thought translates a natural-language problem into a
symbolic reasoning chain, then uses a deterministic solver to execute that
chain and produce the answer \citep{lyu2023faithfulcot}. The answer is thus
generated from the represented computation. \paninidr{} similarly executes
defined constraints over admitted inputs, but also tracks how those inputs
were researched and reviewed across calls. Its workspace exposes both the
lookup dependencies and the calculation; the evidence and local review
outcomes remain attached to the researched values.

\noindent\textbf{ReWOO and LLMCompiler.}
ReWOO separates planning, tool execution, and answer synthesis
\citep{xu2023rewoo}. Its Planner emits interdependent instructions with
placeholders for evidence; Researchers populate those placeholders, and a
Solver combines the plan and observations. LLMCompiler generates a graph
of function calls, dispatches tasks as their dependencies resolve, and
executes independent calls in parallel \citep{kim2024llmcompiler}. It also
supports replanning from intermediate results. These are direct precedents
for explicit executable dependencies. \paninidr{} attaches a research claim,
evidence bindings, and a local review outcome to the values used in such
dependencies, with an admission policy governing their entry into state.

\noindent\textbf{Reasoning trees and modular agent pipelines.}
RT-RAG decomposes a multi-hop question into an explicit reasoning tree,
choosing among candidate trees by consensus and separating the core query
from its known and unknown entities, then traverses the tree bottom-up with
iterative query rewriting and further decomposition when the accumulated
evidence is insufficient \citep{shi2026rtrag}. MA-RAG assigns planning, step
definition, evidence extraction, and answering to separate agents that
exchange intermediate reasoning, with the extractor filtering retrieved
passages before synthesis \citep{nguyen2025marag}. These methods organize
retrieval around explicit intermediate tasks. \paninidr{} connects those
tasks to evidence-linked findings and an admission policy:
a subquestion carries the claim to be checked, an admitted value enters the
shared state with its evidence and review outcome attached, and a checked
revision can add or reformulate subquestions while specified bindings
survive.

\noindent\textbf{Graph-organized evidence.}
SentGraph builds a hierarchical sentence graph offline, distinguishing
nucleus from satellite sentences and linking topic-level subgraphs through
cross-document entity bridges, then performs graph-guided selection and path
expansion at query time \citep{liang2026sentgraph}. HyCE-RAG represents
entities, relations, and contextual evidence as hyperedges, propagates
confidence over the entity--hyperedge incidence structure, and assembles
ranked evidence chains before generation \citep{an2026hycerag}. These
structures are built over the corpus, and their purpose is to present the
model with a coherent chain in place of flat passages. \paninidr{} instead
structures the subquestions: relations connect plan nodes to given anchors and
to one another, and retrieved text is attached as evidence to subquestions
and proposed values. The admission policy governs which values become
available to subsequent assignments. The two are
complementary, since graph-organized retrieval could supply the passages a
research assignment cites.

\noindent\textbf{Slot-bound retrieval.}
ConRAG combines corpus-side evidence links with query-side slot binding:
intermediate answers fill dependency placeholders for later retrieval,
while relation, entity, and text views contribute to a common evidence
ranking \citep{zhu2026conrag}. \paninidr{} shares the use of intermediate
values to guide subsequent work, but centers the execution protocol on
persistent subquestions with claim-scoped review, bounded admission, and
evidence-linked shared state.

\noindent\textbf{Generative Semantic Workspaces.}
\gsw{} represents episodes through entities, roles, and relations to preserve
context beyond isolated facts \citep{rajesh2026gsw,rajesh2026panini}. \paninidr{} uses this
relational vocabulary to express a query's given anchors and typed blanks
at plan nodes. The research harness supplies the execution behavior: assignments
seek values for research nodes, review assesses proposals against their
claims, and admitted bindings support later research or computation.
This separates the inherited representation from the protocol studied here.

\subsection{Deep-Research Harnesses and Persistent Memory}
\label{app:related-harnesses}

DeepResearcher and WebThinker study agents that conduct extended research
through search and tool interaction
\citep{zheng2025deepresearcher,li2025webthinker}. Maintaining continuity over
these interactions is a broader memory problem: MemGPT manages information
across memory tiers, Generative Agents retrieves and reflects on accumulated
experience, and ReSum and context folding compress long interaction contexts
\citep{packer2023memgpt,park2023generativeagents,wu2025resum,sun2025contextfolding}.
The following harnesses make different choices about the form and execution
role of that retained information. We describe their published mechanisms;
Appendices~\ref{app:protocol} and~\ref{app:compat} specify the implementations
and settings evaluated in this paper.

\noindent\textbf{Memory-aware retrieval.}
MIND uses token entropy and attention to trigger retrieval, filters entities
for refined subqueries, and retains retrieved facts across reasoning steps
\citep{ji2025mind}. Its focus on retrieval timing and memory selection is
complementary to \paninidr{}'s subquestion-level interface, which connects
an assignment, a proposed value's evidence and review, and its subsequent
use in the investigation.

\noindent\textbf{SLEUTH.}
SLEUTH maintains source-linked confirmed facts, ranked answer hypotheses,
and prioritized open questions \citep{liu2026sleuth}. Hypotheses reference
supporting and contradicting facts, and the agent rewrites the memory after
tool calls to guide its next action. Runtime checks can enforce the memory
format, while a budget-dependent commitment mechanism encourages synthesis.
\paninidr{} also uses persistent state to guide research, but Researchers submit
local candidate values and evidence for typed subquestions. The harness's
review and admission policy determines how these proposals become bindings
available to subsequent assignments.

\noindent\textbf{Omni-Decision.}
Omni-Decision initializes query-specific evidence needs and fact or
computation slots, tracks their dependencies, and combines critic judgments
with a deterministic reducer that updates shared state
\citep{ma2026omnidecision}. Its planner and critics use a state digest;
dependency completion determines which needs become actionable.
It is therefore a close evidence-state approach, including both explicit
evidence needs and controlled updates. \paninidr{} realizes a particular
combination of subquestions represented through \gsw{}, dispatched research assignments,
review scoped to each candidate's claim and evidence, bounded admission with
recorded status, and explicit binding-reset rules when the plan changes.

\noindent\textbf{HyMem.}
HyMem separates a persistent planner from temporary execution and isolated
reasoning contexts \citep{wang2026hymem}. Structured returns carry findings,
evidence, sources, status, and remaining gaps; episode, working, and tool
memories preserve selected information across context boundaries. Folding
and selective memory injection limit what each role receives.
\paninidr{} likewise separates local work from shared state. Its focus is
the research subquestion: scoped review and admission determine whether
a proposed value becomes an evidence-bound input to later work.

\noindent\textbf{Open Deep Search.}
Open Deep Search combines an Open Search Tool with an agent that decides
how to use search and other tools \citep{alzubi2025opendeepsearch}. The search
component rephrases queries, processes search-result metadata, and scrapes,
chunks, and reranks linked pages. The paper describes ReAct and CodeAct
reasoning-agent variants. \paninidr{} places a subquestion-level execution
protocol around retrieval: assignments specify what to establish, evidence
is attached to relations and proposed values, and admitted bindings provide
inputs for further work.

\noindent\textbf{PyRAG.}
PyRAG decomposes a question into subqueries and synthesizes an executable
Python program over retrieval and answering operations \citep{sun2026pyrag}.
Intermediate answers occupy variables with explicit data dependencies, and
execution records queries, documents, and intermediate results. Runtime
errors can trigger program repair, while insufficient-answer signals can
expand retrieval for affected steps. \paninidr{} also exposes dependencies,
but represents them in a relational research plan linked to candidate review
and admitted evidence bindings. Its revision protocol specifies which
bindings are reset or retained when the plan's subquestions change.

\noindent\textbf{IterResearch.}
IterResearch reconstructs its working context from the original question,
an evolving report, and the preceding action and tool response
\citep{chen2025iterresearch}. Each interaction produces the next report
alongside reasoning and an action, preserving selected findings while
replacing older interaction history. \paninidr{} instead retains research
subquestions and admitted values with evidence and local review status.
Both preserve task-relevant information across calls, but the evolving
report and the subquestion-bound state provide different interfaces for
choosing and carrying out subsequent work.

\noindent\textbf{Re-TRAC.}
Re-TRAC compresses completed research rollouts into structured state for
subsequent rollouts \citep{zhu2026retrac}. Its compression retains partial
answers and conclusions, evidence provenance and source-verification
information, unresolved hypotheses, unexplored branches, and failed attempts.
This memory supports exploration across attempts while reducing repeated
work. \paninidr{} maintains subquestions and bindings within the
investigation, using admitted values to enable dependent assignments and
recording local review at admission. This contrasts rollout-level exploration
memory with bindings that guide individual research assignments.

\noindent\textbf{Answer completion.}
Harnesses differ in how they finish an investigation. SLEUTH uses a
commitment nudge to synthesize from accumulated facts
\citep{liu2026sleuth}, while Omni-Decision can terminate with an
insufficient status when no remaining action can reduce an open need
\citep{ma2026omnidecision}. \paninidr{} recovers an answer from retained
articles when the loop ends without an admitted target, leaving the
workspace unchanged. These policies govern how accumulated research
becomes a delivered answer; Appendix~\ref{app:recovery-analysis}
measures recovery's contribution.

\subsection{Verification, Feedback, and Corrective Retrieval}
\label{app:related-verification}

\noindent\textbf{Learned verifiers and execution feedback.}
\citet{cobbe2021verifiers} train verifiers on generated mathematical solutions
labeled by final-answer correctness and use their scores to select among
sampled solutions. Process supervision instead supplies judgments of
intermediate steps \citep{lightman2023verifystep}. CodeRL uses a critic and
execution feedback to improve code generation \citep{le2022coderl}, while
verifier-guided search studies how to allocate test-time computation using
such evaluations \citep{snell2024scaling}. These approaches use verification
to guide solution selection, generation, or further search.

\noindent\textbf{Iterative feedback and separated checking.}
Self-Refine uses the same underlying LLM to generate an output, give feedback,
and revise it iteratively without additional training
\citep{madaan2023selfrefine}. Reflexion retains verbal feedback as memory for
later attempts \citep{shinn2023reflexion}. Chain-of-Verification first drafts
an answer, generates checking questions, answers those questions, and revises
the draft \citep{dhuliawala2024cove}. Its factored variant isolates each
verification question from the draft and other verification answers.
These are precedents for feedback-driven revision and scoped checking.

\noindent\textbf{LLM-as-a-Verifier.}
\citet{kwok2026llmverifier} develop fine-grained scoring from scoring-token
probabilities, repeated evaluation, and decomposed criteria for assessing
agentic solutions without additional training. This work develops the
checking capability itself; \paninidr{} studies its role in a research
protocol rather than adopting that scoring algorithm. Analyses of agent failures further
motivate tracking how a local decision affects subsequent actions
\citep{zhu2025agentfailures}.

\noindent\textbf{Self-RAG and CRAG.}
Self-RAG trains a model to decide when to retrieve and to generate reflection
tokens assessing passage relevance, support for generated text, and response
utility \citep{asai2024selfrag}. These signals guide retrieval and the
selection of generated segments. CRAG evaluates the relevance of retrieved
documents, then uses the resulting confidence to refine retrieved content,
seek additional web evidence, or combine both \citep{yan2024crag}.
Both use checking to guide retrieval-augmented generation.

\noindent\textbf{Sufficiency and calibrated abstention.}
SURE-RAG treats evidence sufficiency as a property of the retrieved set
rather than of individual passages: a shared claim--evidence verifier scores
each pair, and the local distributions are aggregated into coverage,
relation-strength, uncertainty, and retrieval features that yield a support,
refute, or insufficient decision, with the system answering only when support
is established \citep{qiu2026surerag}. R2VC separates retrieval, generation,
verification, and calibration, using an external natural-language-inference
model to select among structured verdict candidates and a sequence-level
calibrator for selective abstention \citep{dixit2026r2vc}. In both, the
decision governs the final answer, which is either produced or withheld.
\paninidr{} applies its check to an intermediate value instead: a negative
verdict returns the proposal to the Researcher under the shared review
cap rather than ending the question. The admission policy retains the
review outcome in the binding record while making the admitted value
available to later work. A
calibrated sufficiency score could inform that admission decision, which
currently rests on a structured verdict carrying no confidence estimate.

\noindent\textbf{Dependency-guided formal verification.}
The reported formalization of Fermat's Last Theorem illustrates how an
explicit dependency structure can organize work around local checks:
Prove2Me coordinates parallel proof tasks through a dependency graph, while
Lean checks the resulting formal proofs
\citep{anthropic2026fermat,chen2026prove2me}. The relevant connection is the
coordination of dependent assignments and their checking outcomes. Deep
research does not have an equivalent formal proof checker for open-ended
source claims; \paninidr{} instead records scoped model judgments and the
harness's admission decisions alongside the evidence.

Across these lines of work, feedback affects which results are selected,
revised, or supported through further retrieval. \paninidr{} connects these
decisions to the research state. The Verifier assesses a candidate value
against its persistent research subquestion's local claim and evidence;
feedback can prompt further research, while
the harness separately applies its admission policy and preserves the outcome
in the binding record. The Orchestrator can request a plan revision to
reopen a subquestion for further investigation.

\subsection{Attribution and Evidence-Based Evaluation}
\label{app:related-attribution}

Attributable-to-Identified-Sources evaluation asks whether a generated
statement is supported by identified material \citep{rashkin2023attribution}.
ALCE evaluates cited generation, including citation correctness and
completeness \citep{gao2023alce}, and ExpertQA studies attributed answers to
expert-authored questions \citep{malaviya2024expertqa}.
These approaches assess semantic support. Our observed-support measure
instead retains official answer credit when the answer is recoverable from
the available execution evidence record. For computed answers, Obs-D
additionally checks cited inputs and replays the recorded calculation.
The measures thus address evidence availability and recorded computation
alongside, rather than in place of, question-conditioned attribution.

\noindent\textbf{Provenance as a training signal.}
STAMP uses provenance during training rather than evaluation. A
reference-based verifier checks whether each cited document supports an
entity or relation in a training-time evidence graph, first-exposure
attribution traces a supported citation back to the action that first
surfaced it, and the resulting step-level credit is applied by advantage
modulation that leaves the trajectory-level reward unchanged
\citep{xu2026stamp}. \paninidr{} is training-free and uses its records for
measurement and control rather than for credit assignment. The
correspondence is nonetheless direct: an admitted binding already names the
proposal, the evidence attached to it, and the review outcome that admitted
it, which is the provenance a signal of this kind requires.

%% file: main.bbl
\begin{thebibliography}{61}
\providecommand{\natexlab}[1]{#1}
\providecommand{\url}[1]{\texttt{#1}}
\expandafter\ifx\csname urlstyle\endcsname\relax
  \providecommand{\doi}[1]{doi: #1}\else
  \providecommand{\doi}{doi: \begingroup \urlstyle{rm}\Url}\fi

\bibitem[Alzubi et~al.(2025)Alzubi, Brooks, Chiniya, Contente, von Gerlach,
  Irwin, Jiang, Kaz, Nguyen, Oh, Tyagi, and
  Viswanath]{alzubi2025opendeepsearch}
Salaheddin Alzubi, Creston Brooks, Purva Chiniya, Edoardo Contente, Chiara von
  Gerlach, Lucas Irwin, Yihan Jiang, Arda Kaz, Windsor Nguyen, Sewoong Oh,
  Himanshu Tyagi, and Pramod Viswanath.
\newblock Open deep search: Democratizing search with open-source reasoning
  agents, 2025.
\newblock URL \url{https://arxiv.org/abs/2503.20201}.

\bibitem[An et~al.(2026)An, Zhang, Liang, Zhang, Ding, Wu, Li, Su, {Saifullah},
  and Wang]{an2026hycerag}
Hong-Yu An, Yun-Jian Zhang, Chen-Wei Liang, Tian-Yi Zhang, Jian Ding, Yi-Lun
  Wu, Ao-Bo Li, Wei-Cong Su, {Saifullah}, and Mujiangshan Wang.
\newblock {HyCE-RAG}: Hypergraph chain-of-evidence retrieval-augmented
  generation for explainable multi-hop question answering, 2026.
\newblock URL \url{https://arxiv.org/abs/2607.22597}.

\bibitem[{Anthropic}(2026)]{anthropic2026fermat}
{Anthropic}.
\newblock Formalizing {Fermat}'s last theorem.
\newblock Anthropic Research, 2026.
\newblock URL
  \url{https://www.anthropic.com/research/formalizing-fermats-last-theorem}.
\newblock September 4.

\bibitem[Asai et~al.(2024)Asai, Wu, Wang, Sil, and Hajishirzi]{asai2024selfrag}
Akari Asai, Zeqiu Wu, Yizhong Wang, Avirup Sil, and Hannaneh Hajishirzi.
\newblock {Self-RAG}: Learning to retrieve, generate, and critique through
  self-reflection.
\newblock In \emph{International Conference on Learning Representations}, 2024.
\newblock URL \url{https://openreview.net/forum?id=hSyW5go0v8}.

\bibitem[Besta et~al.(2024)Besta, Blach, Kubicek, Gerstenberger, Podstawski,
  Gianinazzi, Gajda, Lehmann, Niewiadomski, Nyczyk, and
  Hoefler]{besta2024graphofthoughts}
Maciej Besta, Nils Blach, Ales Kubicek, Robert Gerstenberger, Michal
  Podstawski, Lukas Gianinazzi, Joanna Gajda, Tomasz Lehmann, Hubert
  Niewiadomski, Piotr Nyczyk, and Torsten Hoefler.
\newblock Graph of thoughts: Solving elaborate problems with large language
  models.
\newblock In \emph{Proceedings of the AAAI Conference on Artificial
  Intelligence}, volume~38, pp.\  17682--17690, 2024.
\newblock \doi{10.1609/aaai.v38i16.29720}.
\newblock URL \url{https://ojs.aaai.org/index.php/AAAI/article/view/29720}.

\bibitem[Chabra et~al.(2026)Chabra, Ziems, Jiang, Khattab, and
  Balakrishnan]{chabra2026dossier}
Om~Chabra, Noah Ziems, Meng Jiang, Omar Khattab, and Hari Balakrishnan.
\newblock Dossier: Deep research via ledger-driven branching search and query
  encoding learning.
\newblock In \emph{Proceedings of the ACM Conference on AI and Agentic
  Systems}, pp.\  263--275. Association for Computing Machinery, 2026.
\newblock \doi{10.1145/3786335.3813122}.
\newblock URL \url{https://people.csail.mit.edu/hari/papers/p193.pdf}.

\bibitem[Chen et~al.(2025{\natexlab{a}})Chen, Qiao, Chen, Yu, Xu, Zhao, Song,
  Yin, Yin, Zhang, Li, Liao, Jiang, Xie, Huang, and Zhou]{chen2025iterresearch}
Guoxin Chen, Zile Qiao, Xuanzhong Chen, Donglei Yu, Haotian Xu, Wayne~Xin Zhao,
  Ruihua Song, Wenbiao Yin, Huifeng Yin, Liwen Zhang, Kuan Li, Minpeng Liao,
  Yong Jiang, Pengjun Xie, Fei Huang, and Jingren Zhou.
\newblock {IterResearch}: Rethinking long-horizon agents with interaction
  scaling, 2025{\natexlab{a}}.
\newblock URL \url{https://arxiv.org/abs/2511.07327}.

\bibitem[Chen et~al.(2026)Chen, Marwaha, Lu, Yuen, and Peng]{chen2026prove2me}
Shuze Chen, Kunal Marwaha, Xiaoyang Lu, Henry Yuen, and Tianyi Peng.
\newblock {Prove2Me}: An open collaborative platform for scaling math
  formalization.
\newblock \emph{arXiv preprint arXiv:2608.28433}, 2026.
\newblock URL \url{https://arxiv.org/abs/2608.28433}.

\bibitem[Chen et~al.(2025{\natexlab{b}})Chen, Ma, Zhuang, Nie, Zou, Liu, Green,
  Patel, Meng, Su, Sharifymoghaddam, Li, Hong, Shi, Liu, Thakur, Zhang, Gao,
  Chen, and Lin]{chen2025browsecompplus}
Zijian Chen, Xueguang Ma, Shengyao Zhuang, Ping Nie, Kai Zou, Andrew Liu,
  Joshua Green, Kshama Patel, Ruoxi Meng, Mingyi Su, Sahel Sharifymoghaddam,
  Yanxi Li, Haoran Hong, Xinyu Shi, Xuye Liu, Nandan Thakur, Crystina Zhang,
  Luyu Gao, Wenhu Chen, and Jimmy Lin.
\newblock Browsecomp-plus: A more fair and transparent evaluation benchmark of
  deep-research agent.
\newblock \emph{arXiv preprint arXiv:2508.06600}, 2025{\natexlab{b}}.

\bibitem[Cobbe et~al.(2021)Cobbe, Kosaraju, Bavarian, Chen, Jun, Kaiser,
  Plappert, Tworek, Hilton, Nakano, Hesse, and Schulman]{cobbe2021verifiers}
Karl Cobbe, Vineet Kosaraju, Mohammad Bavarian, Mark Chen, Heewoo Jun,
  {\L}ukasz Kaiser, Matthias Plappert, Jerry Tworek, Jacob Hilton, Reiichiro
  Nakano, Christopher Hesse, and John Schulman.
\newblock Training verifiers to solve math word problems, 2021.
\newblock URL \url{https://arxiv.org/abs/2110.14168}.

\bibitem[Dhuliawala et~al.(2024)Dhuliawala, Komeili, Xu, Raileanu, Li,
  Celikyilmaz, and Weston]{dhuliawala2024cove}
Shehzaad Dhuliawala, Mojtaba Komeili, Jing Xu, Roberta Raileanu, Xian Li, Asli
  Celikyilmaz, and Jason Weston.
\newblock Chain-of-verification reduces hallucination in large language models.
\newblock In \emph{Findings of the Association for Computational Linguistics:
  ACL 2024}, pp.\  3563--3578. Association for Computational Linguistics, 2024.
\newblock \doi{10.18653/v1/2024.findings-acl.212}.
\newblock URL \url{https://aclanthology.org/2024.findings-acl.212/}.

\bibitem[Dixit \& Pandey(2026)Dixit and Pandey]{dixit2026r2vc}
Dhruv Dixit and Paritosh Pandey.
\newblock {R2VC}: Modular fact-checking with retrieval, verification, and
  confidence calibration, 2026.
\newblock URL \url{https://arxiv.org/abs/2609.11955}.

\bibitem[Gao et~al.(2023)Gao, Yen, Yu, and Chen]{gao2023alce}
Tianyu Gao, Howard Yen, Jiatong Yu, and Danqi Chen.
\newblock Enabling large language models to generate text with citations.
\newblock In \emph{Proceedings of the 2023 Conference on Empirical Methods in
  Natural Language Processing}, pp.\  6465--6488, 2023.
\newblock \doi{10.18653/v1/2023.emnlp-main.398}.
\newblock URL \url{https://aclanthology.org/2023.emnlp-main.398/}.

\bibitem[Hong et~al.(2025)Hong, Troynikov, and Huber]{hong2025contextrot}
Kelly Hong, Anton Troynikov, and Jeff Huber.
\newblock Context rot: How increasing input tokens impacts {LLM} performance.
\newblock Technical report, Chroma, July 2025.
\newblock URL \url{https://trychroma.com/research/context-rot}.

\bibitem[Izacard \& Grave(2021)Izacard and Grave]{izacard2021leveraging}
Gautier Izacard and Edouard Grave.
\newblock Leveraging passage retrieval with generative models for open domain
  question answering.
\newblock In \emph{Proceedings of the 16th Conference of the European Chapter
  of the Association for Computational Linguistics: Main Volume}, pp.\
  874--880, 2021.
\newblock \doi{10.18653/v1/2021.eacl-main.74}.
\newblock URL \url{https://aclanthology.org/2021.eacl-main.74/}.

\bibitem[Ji et~al.(2025)Ji, Meng, Li, and He]{ji2025mind}
Yuelyu Ji, Rui Meng, Zhuochun Li, and Daqing He.
\newblock Memory-aware and uncertainty-guided retrieval for multi-hop question
  answering, 2025.
\newblock URL \url{https://arxiv.org/abs/2503.23095}.

\bibitem[Kim et~al.(2024)Kim, Moon, Tabrizi, Lee, Mahoney, Keutzer, and
  Gholami]{kim2024llmcompiler}
Sehoon Kim, Suhong Moon, Ryan Tabrizi, Nicholas Lee, Michael~W. Mahoney, Kurt
  Keutzer, and Amir Gholami.
\newblock An {LLM} compiler for parallel function calling.
\newblock In \emph{Proceedings of the 41st International Conference on Machine
  Learning}, volume 235, pp.\  24370--24391. PMLR, 2024.
\newblock URL \url{https://proceedings.mlr.press/v235/kim24y.html}.

\bibitem[Krishna et~al.(2024)Krishna, Krishna, Mohananey, Schwarcz, Stambler,
  Upadhyay, and Faruqui]{krishna2024frames}
Satyapriya Krishna, Kalpesh Krishna, Anhad Mohananey, Steven Schwarcz, Adam
  Stambler, Shyam Upadhyay, and Manaal Faruqui.
\newblock Fact, fetch, and reason: A unified evaluation of retrieval-augmented
  generation, 2024.
\newblock URL \url{https://arxiv.org/abs/2409.12941}.

\bibitem[Kwok et~al.(2026)Kwok, Li, Atreya, Liu, Jiang, Finn, Pavone, Stoica,
  and Mirhoseini]{kwok2026llmverifier}
Jacky Kwok, Shulu Li, Pranav Atreya, Yuejiang Liu, Yixing Jiang, Chelsea Finn,
  Marco Pavone, Ion Stoica, and Azalia Mirhoseini.
\newblock {LLM}-as-a-verifier: A general-purpose verification framework.
\newblock \emph{arXiv preprint arXiv:2607.05391}, 2026.
\newblock URL \url{https://arxiv.org/abs/2607.05391}.

\bibitem[Le et~al.(2022)Le, Wang, Gotmare, Savarese, and Hoi]{le2022coderl}
Hung Le, Yue Wang, Akhilesh~Deepak Gotmare, Silvio Savarese, and Steven C.~H.
  Hoi.
\newblock {CodeRL}: Mastering code generation through pretrained models and
  deep reinforcement learning.
\newblock In \emph{Advances in Neural Information Processing Systems},
  volume~35, pp.\  21314--21328, 2022.
\newblock URL \url{https://arxiv.org/abs/2207.01780}.

\bibitem[Lewis et~al.(2020)Lewis, Perez, Piktus, Petroni, Karpukhin, Goyal,
  K{\"u}ttler, Lewis, Yih, Rockt{\"a}schel, Riedel, and Kiela]{lewis2020rag}
Patrick Lewis, Ethan Perez, Aleksandra Piktus, Fabio Petroni, Vladimir
  Karpukhin, Naman Goyal, Heinrich K{\"u}ttler, Mike Lewis, Wen-tau Yih, Tim
  Rockt{\"a}schel, Sebastian Riedel, and Douwe Kiela.
\newblock Retrieval-augmented generation for knowledge-intensive {NLP} tasks.
\newblock In \emph{Advances in Neural Information Processing Systems},
  volume~33, 2020.
\newblock URL
  \url{https://papers.neurips.cc/paper/2020/hash/6b493230205f780e1bc26945df7481e5-Abstract.html}.

\bibitem[Li et~al.(2025)Li, Jin, Dong, Qian, Wu, Wen, Zhu, and
  Dou]{li2025webthinker}
Xiaoxi Li, Jiajie Jin, Guanting Dong, Hongjin Qian, Yongkang Wu, Ji-Rong Wen,
  Yutao Zhu, and Zhicheng Dou.
\newblock {WebThinker}: Empowering large reasoning models with deep research
  capability.
\newblock In \emph{Advances in Neural Information Processing Systems},
  volume~38, 2025.
\newblock \doi{10.52202/085713-4011}.
\newblock URL
  \url{https://papers.nips.cc/paper_files/paper/2025/hash/ae03bdef276132fae089692445725635-Abstract-Conference.html}.

\bibitem[Liang et~al.(2026)Liang, Zhou, Zhou, Qing, Zhao, Wang, Song, and
  Li]{liang2026sentgraph}
Junli Liang, Pengfei Zhou, Wangqiu Zhou, Wenjie Qing, Qi~Zhao, Ziwen Wang,
  Qi~Song, and Xiangyang Li.
\newblock {SentGraph}: Hierarchical sentence graph for multi-hop
  retrieval-augmented question answering, 2026.
\newblock URL \url{https://arxiv.org/abs/2601.03014}.

\bibitem[Lightman et~al.(2023)Lightman, Kosaraju, Burda, Edwards, Baker, Lee,
  Leike, Schulman, Sutskever, and Cobbe]{lightman2023verifystep}
Hunter Lightman, Vineet Kosaraju, Yura Burda, Harri Edwards, Bowen Baker, Teddy
  Lee, Jan Leike, John Schulman, Ilya Sutskever, and Karl Cobbe.
\newblock Let's verify step by step.
\newblock \emph{arXiv preprint arXiv:2305.20050}, 2023.
\newblock URL \url{https://arxiv.org/abs/2305.20050}.

\bibitem[Liu et~al.(2024)Liu, Lin, Hewitt, Paranjape, Bevilacqua, Petroni, and
  Liang]{liu2024lostmiddle}
Nelson~F. Liu, Kevin Lin, John Hewitt, Ashwin Paranjape, Michele Bevilacqua,
  Fabio Petroni, and Percy Liang.
\newblock Lost in the middle: How language models use long contexts.
\newblock \emph{Transactions of the Association for Computational Linguistics},
  12:\penalty0 157--173, 2024.
\newblock \doi{10.1162/tacl_a_00638}.
\newblock URL \url{https://aclanthology.org/2024.tacl-1.9/}.

\bibitem[Liu(2026)]{liu2026sleuth}
Ning Liu.
\newblock Track, rank, crack: Epistemic working memory scales multi-hop
  reasoning in language agents, 2026.
\newblock URL \url{https://arxiv.org/abs/2607.12267}.

\bibitem[Lyu et~al.(2023)Lyu, Havaldar, Stein, Zhang, Rao, Wong, Apidianaki,
  and Callison-Burch]{lyu2023faithfulcot}
Qing Lyu, Shreya Havaldar, Adam Stein, Li~Zhang, Delip Rao, Eric Wong, Marianna
  Apidianaki, and Chris Callison-Burch.
\newblock Faithful chain-of-thought reasoning.
\newblock In \emph{Proceedings of the 13th International Joint Conference on
  Natural Language Processing and the 3rd Conference of the Asia-Pacific
  Chapter of the Association for Computational Linguistics (Volume 1: Long
  Papers)}, pp.\  305--329. Association for Computational Linguistics, 2023.
\newblock \doi{10.18653/v1/2023.ijcnlp-main.20}.
\newblock URL \url{https://aclanthology.org/2023.ijcnlp-main.20/}.

\bibitem[Ma et~al.(2026)Ma, Zhu, Zhong, Zhu, Sun, Shi, Mei, Yang, and
  Hoi]{ma2026omnidecision}
Ming Ma, Yi~Zhu, Yiran Zhong, Feida Zhu, Weigao Sun, Junhan Shi, Lingrui Mei,
  Tianming Yang, and Steven Hoi.
\newblock {Omni-Decision}: A progressive evidence-state agent system for
  omni-modal {QA}, 2026.
\newblock URL \url{https://arxiv.org/abs/2607.11433}.

\bibitem[Madaan et~al.(2023)Madaan, Tandon, Gupta, Hallinan, Gao, Wiegreffe,
  Alon, Dziri, Prabhumoye, Yang, Gupta, Majumder, Hermann, Welleck,
  Yazdanbakhsh, and Clark]{madaan2023selfrefine}
Aman Madaan, Niket Tandon, Prakhar Gupta, Skyler Hallinan, Luyu Gao, Sarah
  Wiegreffe, Uri Alon, Nouha Dziri, Shrimai Prabhumoye, Yiming Yang, Shashank
  Gupta, Bodhisattwa~Prasad Majumder, Katherine Hermann, Sean Welleck, Amir
  Yazdanbakhsh, and Peter Clark.
\newblock {Self-Refine}: Iterative refinement with self-feedback.
\newblock In \emph{Advances in Neural Information Processing Systems},
  volume~36, pp.\  46534--46594, 2023.
\newblock URL \url{https://arxiv.org/abs/2303.17651}.

\bibitem[Malaviya et~al.(2024)Malaviya, Lee, Chen, Sieber, Yatskar, and
  Roth]{malaviya2024expertqa}
Chaitanya Malaviya, Subin Lee, Sihao Chen, Elizabeth Sieber, Mark Yatskar, and
  Dan Roth.
\newblock {ExpertQA}: Expert-curated questions and attributed answers.
\newblock In \emph{Proceedings of the 2024 Conference of the North American
  Chapter of the Association for Computational Linguistics: Human Language
  Technologies (Volume 1: Long Papers)}, pp.\  3025--3045, 2024.
\newblock \doi{10.18653/v1/2024.naacl-long.167}.
\newblock URL \url{https://aclanthology.org/2024.naacl-long.167/}.

\bibitem[Nguyen et~al.(2025)Nguyen, Chin, and Tai]{nguyen2025marag}
Thang Nguyen, Peter Chin, and Yu-Wing Tai.
\newblock {MA-RAG}: Multi-agent retrieval-augmented generation via
  collaborative chain-of-thought reasoning, 2025.
\newblock URL \url{https://arxiv.org/abs/2505.20096}.

\bibitem[{OpenAI}(2025)]{openai2025gptoss}
{OpenAI}.
\newblock {gpt-oss-120b} \& {gpt-oss-20b} model card, 2025.
\newblock URL \url{https://openai.com/index/gpt-oss-model-card/}.

\bibitem[Packer et~al.(2023)Packer, Wooders, Lin, Fang, Patil, Stoica, and
  Gonzalez]{packer2023memgpt}
Charles Packer, Sarah Wooders, Kevin Lin, Vivian Fang, Shishir~G. Patil, Ion
  Stoica, and Joseph~E. Gonzalez.
\newblock {MemGPT}: Towards {LLM}s as operating systems.
\newblock \emph{arXiv preprint arXiv:2310.08560}, 2023.
\newblock URL \url{https://arxiv.org/abs/2310.08560}.

\bibitem[Park et~al.(2023)Park, O'Brien, Cai, Morris, Liang, and
  Bernstein]{park2023generativeagents}
Joon~Sung Park, Joseph O'Brien, Carrie~Jun Cai, Meredith~Ringel Morris, Percy
  Liang, and Michael~S. Bernstein.
\newblock Generative agents: Interactive simulacra of human behavior.
\newblock In \emph{Proceedings of the 36th Annual ACM Symposium on User
  Interface Software and Technology}, 2023.
\newblock \doi{10.1145/3586183.3606763}.
\newblock URL \url{https://doi.org/10.1145/3586183.3606763}.

\bibitem[Press et~al.(2023)Press, Zhang, Min, Schmidt, Smith, and
  Lewis]{press2023selfask}
Ofir Press, Muru Zhang, Sewon Min, Ludwig Schmidt, Noah Smith, and Mike Lewis.
\newblock Measuring and narrowing the compositionality gap in language models.
\newblock In \emph{Findings of the Association for Computational Linguistics:
  EMNLP 2023}, pp.\  5687--5711. Association for Computational Linguistics,
  2023.
\newblock \doi{10.18653/v1/2023.findings-emnlp.378}.
\newblock URL \url{https://aclanthology.org/2023.findings-emnlp.378/}.

\bibitem[Qiu et~al.(2026)Qiu, Han, and Huang]{qiu2026surerag}
Jingxi Qiu, Zeyu Han, and Cheng Huang.
\newblock {SURE-RAG}: Sufficiency and uncertainty-aware evidence verification
  for selective retrieval-augmented generation, 2026.
\newblock URL \url{https://arxiv.org/abs/2605.03534}.

\bibitem[Rajesh et~al.(2026{\natexlab{a}})Rajesh, Holur, Duan, Chong, and
  Roychowdhury]{rajesh2026gsw}
Shreyas Rajesh, Pavan Holur, Chenda Duan, David Chong, and Vwani Roychowdhury.
\newblock Beyond fact retrieval: Episodic memory for {RAG} with generative
  semantic workspaces.
\newblock In \emph{Proceedings of the AAAI Conference on Artificial
  Intelligence}, volume~40, pp.\  32782--32790, 2026{\natexlab{a}}.
\newblock \doi{10.1609/aaai.v40i39.40557}.
\newblock URL \url{https://ojs.aaai.org/index.php/AAAI/article/view/40557}.

\bibitem[Rajesh et~al.(2026{\natexlab{b}})Rajesh, Holur, Turali, Duan, and
  Roychowdhury]{rajesh2026panini}
Shreyas Rajesh, Pavan Holur, Mehmet~Yigit Turali, Chenda Duan, and Vwani
  Roychowdhury.
\newblock {Panini}: Continual learning in token space via structured memory,
  2026{\natexlab{b}}.
\newblock URL \url{https://arxiv.org/abs/2602.15156}.

\bibitem[Rashkin et~al.(2023)Rashkin, Nikolaev, Lamm, Aroyo, Collins, Das,
  Petrov, Tomar, Turc, and Reitter]{rashkin2023attribution}
Hannah Rashkin, Vitaly Nikolaev, Matthew Lamm, Lora Aroyo, Michael Collins,
  Dipanjan Das, Slav Petrov, Gaurav~Singh Tomar, Iulia Turc, and David Reitter.
\newblock Measuring attribution in natural language generation models.
\newblock \emph{Computational Linguistics}, 49\penalty0 (4):\penalty0 777--840,
  2023.
\newblock \doi{10.1162/coli_a_00486}.
\newblock URL \url{https://aclanthology.org/2023.cl-4.2/}.

\bibitem[Shi et~al.(2026)Shi, Sun, Liu, Yang, Fang, Sun, and Gu]{shi2026rtrag}
Yuling Shi, Maolin Sun, Zijun Liu, Mo~Yang, Yixiong Fang, Tianran Sun, and
  Xiaodong Gu.
\newblock Reasoning in trees: Improving retrieval-augmented generation for
  multi-hop question answering, 2026.
\newblock URL \url{https://arxiv.org/abs/2601.11255}.

\bibitem[Shinn et~al.(2023)Shinn, Cassano, Gopinath, Narasimhan, and
  Yao]{shinn2023reflexion}
Noah Shinn, Federico Cassano, Ashwin Gopinath, Karthik Narasimhan, and Shunyu
  Yao.
\newblock Reflexion: Language agents with verbal reinforcement learning.
\newblock In \emph{Advances in Neural Information Processing Systems},
  volume~36, 2023.
\newblock URL
  \url{https://proceedings.neurips.cc/paper_files/paper/2023/hash/1b44b878bb782e6954cd888628510e90-Abstract-Conference.html}.

\bibitem[Snell et~al.(2024)Snell, Lee, Xu, and Kumar]{snell2024scaling}
Charlie Snell, Jaehoon Lee, Kelvin Xu, and Aviral Kumar.
\newblock Scaling {LLM} test-time compute optimally can be more effective than
  scaling model parameters.
\newblock \emph{arXiv preprint arXiv:2408.03314}, 2024.
\newblock URL \url{https://arxiv.org/abs/2408.03314}.

\bibitem[Sun et~al.(2026)Sun, Shi, Xie, Wang, Parekh, Jiang, Shi, Fan, Zheng,
  Li, Wang, Liu, and Han]{sun2026pyrag}
Jiashuo Sun, Jimeng Shi, Yixuan Xie, Saizhuo Wang, Jash~Rajesh Parekh,
  Pengcheng Jiang, Zhiyi Shi, Jiajun Fan, Qinglong Zheng, Peiran Li, Shaowen
  Wang, Ge~Liu, and Jiawei Han.
\newblock Retrieval is cheap, show me the code: Executable multi-hop reasoning
  for retrieval-augmented generation, 2026.
\newblock URL \url{https://arxiv.org/abs/2605.12975}.

\bibitem[Sun et~al.(2025)Sun, Lu, Ling, Liu, Yao, Yang, and
  Chen]{sun2025contextfolding}
Weiwei Sun, Miao Lu, Zhan Ling, Kang Liu, Xuesong Yao, Yiming Yang, and Jiecao
  Chen.
\newblock Scaling long-horizon llm agent via context-folding, 2025.
\newblock URL \url{https://arxiv.org/abs/2510.11967}.

\bibitem[Trivedi et~al.(2023)Trivedi, Balasubramanian, Khot, and
  Sabharwal]{trivedi2023ircot}
Harsh Trivedi, Niranjan Balasubramanian, Tushar Khot, and Ashish Sabharwal.
\newblock Interleaving retrieval with chain-of-thought reasoning for
  knowledge-intensive multi-step questions.
\newblock In \emph{Proceedings of the 61st Annual Meeting of the Association
  for Computational Linguistics (Volume 1: Long Papers)}, pp.\  10014--10037,
  2023.
\newblock \doi{10.18653/v1/2023.acl-long.557}.
\newblock URL \url{https://aclanthology.org/2023.acl-long.557/}.

\bibitem[Wan et~al.(2026)Wan, Fang, Li, Huo, Wang, Mi, Yu, and
  Lyu]{wan2026deepverifier}
Yuxuan Wan, Tianqing Fang, Zaitang Li, Yintong Huo, Wenxuan Wang, Haitao Mi,
  Dong Yu, and Michael~R. Lyu.
\newblock Inference-time scaling of verification: Self-evolving deep research
  agents via test-time rubric-guided verification.
\newblock In \emph{Findings of the Association for Computational Linguistics:
  ACL 2026}, pp.\  24822--24835. Association for Computational Linguistics,
  2026.
\newblock \doi{10.18653/v1/2026.findings-acl.1243}.
\newblock URL \url{https://aclanthology.org/2026.findings-acl.1243/}.

\bibitem[Wang et~al.(2026)Wang, Xiao, Cui, Zhang, Wang, Zhang, and
  Xu]{wang2026hymem}
XinQi Wang, Jinwei Xiao, Sijia Cui, Hongming Zhang, Yanna Wang, Qingyang Zhang,
  and Bo~Xu.
\newblock {HyMem}: Hierarchical context management for long-horizon agents via
  information isolation.
\newblock \emph{arXiv preprint arXiv:2608.15703}, 2026.
\newblock URL \url{https://arxiv.org/abs/2608.15703}.

\bibitem[Wang et~al.(2023)Wang, Wei, Schuurmans, Le, Chi, Narang, Chowdhery,
  and Zhou]{wang2023selfconsistency}
Xuezhi Wang, Jason Wei, Dale Schuurmans, Quoc Le, Ed~Chi, Sharan Narang,
  Aakanksha Chowdhery, and Denny Zhou.
\newblock Self-consistency improves chain of thought reasoning in language
  models.
\newblock In \emph{International Conference on Learning Representations}, 2023.
\newblock URL \url{https://openreview.net/forum?id=1PL1NIMMrw}.

\bibitem[Wei et~al.(2022)Wei, Wang, Schuurmans, Bosma, Ichter, Xia, Chi, Le,
  and Zhou]{wei2022chainofthought}
Jason Wei, Xuezhi Wang, Dale Schuurmans, Maarten Bosma, Brian Ichter, Fei Xia,
  Ed~H. Chi, Quoc~V. Le, and Denny Zhou.
\newblock Chain-of-thought prompting elicits reasoning in large language
  models.
\newblock In \emph{Advances in Neural Information Processing Systems},
  volume~35, 2022.
\newblock URL
  \url{https://proceedings.neurips.cc/paper_files/paper/2022/hash/9d5609613524ecf4f15af0f7b31abca4-Abstract-Conference.html}.

\bibitem[Wolfson et~al.(2025)Wolfson, Trivedi, Geva, Goldberg, Roth, Khot,
  Sabharwal, and Tsarfaty]{wolfson2025monaco}
Tomer Wolfson, Harsh Trivedi, Mor Geva, Yoav Goldberg, Dan Roth, Tushar Khot,
  Ashish Sabharwal, and Reut Tsarfaty.
\newblock {MoNaCo}: More natural and complex questions for reasoning across
  dozens of documents, 2025.
\newblock URL \url{https://arxiv.org/abs/2508.11133}.

\bibitem[Wu et~al.(2025)Wu, Li, Zhao, Zhang, Ou, Yin, Zhang, Jiang, Xie, Huang,
  Cheng, Wang, Cheng, and Zhou]{wu2025resum}
Xixi Wu, Kuan Li, Yida Zhao, Liwen Zhang, Litu Ou, Huifeng Yin, Zhongwang
  Zhang, Yong Jiang, Pengjun Xie, Fei Huang, Minhao Cheng, Shuai Wang, Hong
  Cheng, and Jingren Zhou.
\newblock {ReSum}: Unlocking long-horizon search intelligence via context
  summarization.
\newblock \emph{arXiv preprint arXiv:2509.13313}, 2025.
\newblock URL \url{https://arxiv.org/abs/2509.13313}.

\bibitem[Xu et~al.(2023)Xu, Peng, Lei, Mukherjee, Liu, and Xu]{xu2023rewoo}
Binfeng Xu, Zhiyuan Peng, Bowen Lei, Subhabrata Mukherjee, Yuchen Liu, and
  Dongkuan Xu.
\newblock {ReWOO}: Decoupling reasoning from observations for efficient
  augmented language models, 2023.
\newblock URL \url{https://arxiv.org/abs/2305.18323}.

\bibitem[Xu et~al.(2026)Xu, Xu, Chen, Wang, Li, Liu, Wu, Xia, and
  Li]{xu2026stamp}
Ke~Xu, Han Xu, Xinran Chen, Yuqian Wang, Zhixuan Li, Xiaojian Liu, Changwo Wu,
  Jianqiang Xia, and Yuchen Li.
\newblock {STAMP}: Provenance-guided credit assignment for deep search agents,
  2026.
\newblock URL \url{https://arxiv.org/abs/2607.11172}.

\bibitem[Yan et~al.(2024)Yan, Gu, Zhu, and Ling]{yan2024crag}
Shi-Qi Yan, Jia-Chen Gu, Yun Zhu, and Zhen-Hua Ling.
\newblock Corrective retrieval augmented generation, 2024.
\newblock URL \url{https://arxiv.org/abs/2401.15884}.

\bibitem[Yao et~al.(2023{\natexlab{a}})Yao, Yu, Zhao, Shafran, Griffiths, Cao,
  and Narasimhan]{yao2023treeofthoughts}
Shunyu Yao, Dian Yu, Jeffrey Zhao, Izhak Shafran, Thomas~L. Griffiths, Yuan
  Cao, and Karthik Narasimhan.
\newblock Tree of thoughts: Deliberate problem solving with large language
  models.
\newblock In \emph{Advances in Neural Information Processing Systems},
  volume~36, 2023{\natexlab{a}}.
\newblock URL
  \url{https://proceedings.neurips.cc/paper_files/paper/2023/hash/271db9922b8d1f4dd7aaef84ed5ac703-Abstract-Conference.html}.

\bibitem[Yao et~al.(2023{\natexlab{b}})Yao, Zhao, Yu, Du, Shafran, Narasimhan,
  and Cao]{yao2023react}
Shunyu Yao, Jeffrey Zhao, Dian Yu, Nan Du, Izhak Shafran, Karthik Narasimhan,
  and Yuan Cao.
\newblock {ReAct}: Synergizing reasoning and acting in language models.
\newblock In \emph{International Conference on Learning Representations},
  2023{\natexlab{b}}.
\newblock URL \url{https://openreview.net/forum?id=WE_vluYUL-X}.

\bibitem[Zheng et~al.(2025)Zheng, Fu, Hu, Cai, Ye, Lu, and
  Liu]{zheng2025deepresearcher}
Yuxiang Zheng, Dayuan Fu, Xiangkun Hu, Xiaojie Cai, Lyumanshan Ye, Pengrui Lu,
  and Pengfei Liu.
\newblock {DeepResearcher}: Scaling deep research via reinforcement learning in
  real-world environments.
\newblock In \emph{Proceedings of the 2025 Conference on Empirical Methods in
  Natural Language Processing}, pp.\  414--431, 2025.
\newblock \doi{10.18653/v1/2025.emnlp-main.22}.
\newblock URL \url{https://aclanthology.org/2025.emnlp-main.22/}.

\bibitem[Zhu et~al.(2026{\natexlab{a}})Zhu, Jia, Lan, Ren, Gu, Jiang, Wang, Xu,
  and Luo]{zhu2026marco}
Bin Zhu, Qianghuai Jia, Tian Lan, Junyang Ren, Feng Gu, Feihu Jiang, Longyue
  Wang, Zhao Xu, and Weihua Luo.
\newblock Marco deepresearch: Unlocking efficient deep research agents via
  verification-centric design, 2026{\natexlab{a}}.
\newblock URL \url{https://arxiv.org/abs/2603.28376}.

\bibitem[Zhu et~al.(2026{\natexlab{b}})Zhu, Zhang, Ma, Xu, Zhang, Yang, Wang,
  Qiu, Wu, Dai, Ma, Liu, Yang, Luo, Yang, Li, Wang, Chen, Geng, and
  Guo]{zhu2026retrac}
Jialiang Zhu, Gongrui Zhang, Xiaolong Ma, Lin Xu, Miaosen Zhang, Ruiqi Yang,
  Song Wang, Kai Qiu, Zhirong Wu, Qi~Dai, Ruichun Ma, Bei Liu, Yifan Yang,
  Chong Luo, Zhengyuan Yang, Linjie Li, Lijuan Wang, Weizhu Chen, Xin Geng, and
  Baining Guo.
\newblock {RE-TRAC}: Recursive trajectory compression for deep search agents,
  2026{\natexlab{b}}.
\newblock URL \url{https://arxiv.org/abs/2602.02486}.

\bibitem[Zhu et~al.(2025)Zhu, Liu, Li, Tian, Yang, Zhang, Han, Xie, Cui, Zhang,
  Ma, Yu, Ramesh, Wu, Liu, Lu, Zou, and You]{zhu2025agentfailures}
Kunlun Zhu, Zijia Liu, Bingxuan Li, Muxin Tian, Yingxuan Yang, Jiaxun Zhang,
  Pengrui Han, Qipeng Xie, Fuyang Cui, Weijia Zhang, Xiaoteng Ma, Xiaodong Yu,
  Gowtham Ramesh, Jialian Wu, Zicheng Liu, Pan Lu, James Zou, and Jiaxuan You.
\newblock Where {LLM} agents fail and how they can learn from failures.
\newblock \emph{arXiv preprint arXiv:2509.25370}, 2025.
\newblock URL \url{https://arxiv.org/abs/2509.25370}.

\bibitem[Zhu et~al.(2026{\natexlab{c}})Zhu, Chen, Zhong, Liu, and
  Du]{zhu2026conrag}
Yikai Zhu, Kunfeng Chen, Qihuang Zhong, Juhua Liu, and Bo~Du.
\newblock {ConRAG}: Consensus-driven multi-view retrieval for multi-hop
  question answering, 2026{\natexlab{c}}.
\newblock URL \url{https://arxiv.org/abs/2605.28093}.

\end{thebibliography}
